\pdfoutput=1

\documentclass[11pt]{article}

\usepackage[]{acl}

\usepackage{times}
\usepackage{latexsym}

\usepackage[T1]{fontenc}

\usepackage[utf8]{inputenc}
\usepackage{times}
\usepackage{epsfig}
\usepackage{graphicx}
\usepackage{amsmath,amssymb,amsfonts}
\usepackage{multirow}
\usepackage{makecell}
\usepackage{colortbl}
\usepackage{multirow}
\usepackage{subfigure}
\usepackage{float}
\usepackage{enumitem}
\usepackage{booktabs}
\usepackage{xcolor}

\newenvironment{packed_item}{
\begin{itemize}
  \setlength{\itemsep}{1pt}
  \setlength{\parskip}{0pt}
  \setlength{\parsep}{0pt}
}{\end{itemize}}
\allowdisplaybreaks[4]

\usepackage{microtype}

\title{QLEVR: A Diagnostic Dataset for Quantificational Language and Elementary Visual Reasoning}

\author{Zechen Li \\
  Northeastern University, US \\
  \texttt{li.zec@northeastern.edu} \\\And
  Anders Søgaard \\
  University of Copenhagen, Denmark \\
  \texttt{soegaard@di.ku.dk} \\}
  
\begin{document}
\maketitle

\begin{abstract}
Synthetic datasets have successfully been used to probe visual question-answering datasets for their reasoning abilities. CLEVR \cite{johnson2017clevr}, for example, tests a range of visual reasoning abilities. The questions in CLEVR focus on comparisons of shapes, colors, and sizes, numerical reasoning, and existence claims. This paper introduces a minimally biased, diagnostic visual question-answering dataset, QLEVR, that {\em goes beyond existential and numerical} quantification and focus on more complex quantifiers and their combinations, e.g., asking whether there are {\em more than two red balls that are smaller than at least three blue balls} in an image. We describe how the dataset was created and present a first evaluation of state-of-the-art visual question-answering models, showing that QLEVR presents a formidable challenge to our current models.  Code and Dataset are available at \url{https://github.com/zechenli03/QLEVR}
\end{abstract}

\section{Introduction}
\label{Introduction}

\begin{figure}[t]
  \centering
  \includegraphics[width=0.48\textwidth]{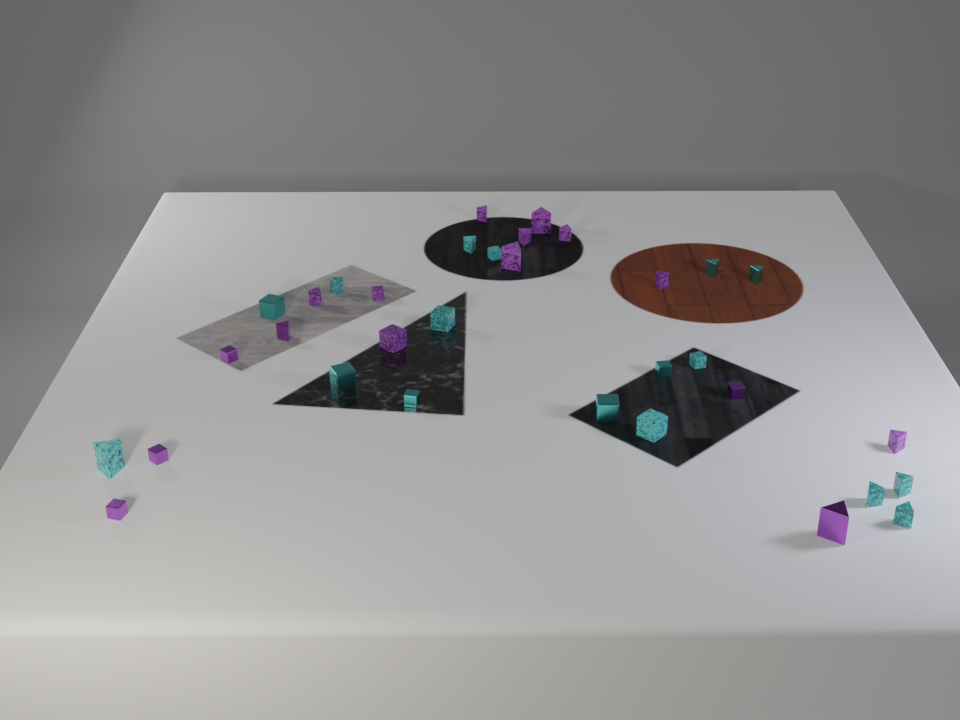} \\
  \vspace{2mm}
  \begin{minipage}{0.48\textwidth}
    \footnotesize
    \textbf{Question:} Are \textcolor{purple}{all} the cyan metallic triangular prisms on the brown plane? \\*
    \textbf{Answer:} True \\*[2mm]
    \textbf{Question:} On the non-white planes on the left rear side of the black wood rectangular plane, \textcolor{purple}{all} the cyan metallic cubes \textcolor{purple}{but at least 2} are larger than \textcolor{purple}{at most 7} cubes; is it right? \\*
    \textbf{Answer:} False
  \end{minipage}
  \caption{A sample image and questions from QLEVR. Tasks involve attribute recognition, counting, comparing numbers, spatial relationships, and understanding of \textcolor{purple}{quantifiers}.}
  \label{fig:intro}
\end{figure}

Visual question answering is at the locus of computer vision and natural language processing, and its objective is developing computer vision systems that can answer arbitrary natural language questions about images \cite{NIPS2016_9dcb88e0,NIPS2017_05192834,NEURIPS2018_67d96d45,NEURIPS2020_20d749bc}. This is useful across a range of applications, including medical image analysis, accessibility for visually impaired, video surveillance, art and advertisement \cite{Barra2021VisualQA}. 

The complexity of visual question answering naturally depends on the complexity of the images and the complexity of the natural language questions. The task reduces to object recognition for very simple questions of the form: 

\begin{itemize}
    \item[(1)] Is there a triangle in this image?
\end{itemize}

Object recognition can of course be a very complex task on its own, depending on the types of objects, the number of possible objects to be recognized, the amount of supervision for inducing a good model, general image quality, etc. However, more complex queries such as (2) make visual question answering much harder: 

\begin{itemize}
    \item[(2)] Is there a triangle inside a circle in this image?
\end{itemize}

Answering such a question in the presence of an image requires a computer vision system that not only recognizes objects, but also relations between them. CLEVR \cite{johnson2017clevr} probes computer vision systems's ability to answer even more complex queries, such as, for instance:

\begin{itemize}
    \item[(3)] Is there a cyan cube to the right of the yellow sphere?
\end{itemize}

Question (3) involves reasoning about the relation between two objects, as well as the compositional semantics of color adjectives. In addition to shapes and colors, CLEVR also includes questions about sizes and quantities. 

In this paper, we present a novel visual question-answering dataset that goes beyond CLEVR in focusing specifically on {\em quantificational language}, e.g.:  

\begin{itemize}
    \item[(4)] Are most of the cyan cubes to the right of the yellow sphere?
\end{itemize}

Given the complexity of quantificational language, the rich typology of expressions of quantification across different languages, and the interest from philosophy, it is perhaps surprising that quantificational language has received relatively little attention in the NLP community (see \S2), but we believe it is a crucial step in pushing the research horizons in (visual) question-answering. 

\paragraph{Contributions} Based on a comprehensive typology of English quantifiers, we build a dataset of 100,000 synthetic images and 999,446 unique questions to these images. This is roughly the same size as or a little bigger than CLEVR \cite{johnson2017clevr}. Our questions are on average longer than previous datasets. We evaluate three baselines from \citet{johnson2017clevr}, a text-only baseline based on BERT \cite{devlin2018bert}, and MAC \cite{hudson2018compositional} on QLEVR and analyze performance across quantifier types.  
\section{Related Work}
\label{Related Work}

\paragraph{Visual Question Answering Challenge Datasets} Several synthetic challenge datasets for visual question answering exist: \citet{7780381} presents SHAPES, a predecessor to CLEVR and QLEVR, relying also on synthetic constellations of colored geometric shapes and template-driven question generation. \citet{pezzelle-fernandez-2019-red} create a similar dataset to probe visual question answering models for knowledge of adjectival semantics. A portion of the visual question answering dataset \cite{10.1007/s11263-016-0966-6} contains synthetic cartoon imagery.  \citet{sampat-etal-2021-clevr} present an extension of CLEVR that probes for hypothetical reasoning of the form: {\em If someone removed three triangles from this image, how many would be left?} \citet{conf/nips/MalinowskiF14} combined natural images with synthetic, template-driven question generation. Finally, \citet{parfenova-etal-2021-probing} recently created a dataset of three-image scenes to probe two-step reasoning. 

Synthetic visual question answering datasets have several advantages over ones based on real images and questions that tend to suffer from selection biases \cite{liu-etal-2021-visually}, but of course they are limited in what can be induced from them. They are therefore mostly useful for probing the limitations of visual question answering architectures and off-the-shelf models. Showing results only on synthetic data is often seen as a weakness in the literature \cite{Hassantabar2018VisualQA}, but synthetic data is useful for diagnosing the errors of visual question answering systems, in our case highlighting the challenges posed by quantifiers.

\paragraph{Quantifiers} Quantifiers have been largely ignored in the NLP community. Question-answering datasets have been developed for numerical reasoning in English \cite{dua-etal-2019-drop}, and some have identified quantifier words as important sources of errors for textual entailment systems \cite{joshi-etal-2020-taxinli}. \citet{Fang2021PartW} recently focused on the two quantifier words {\em part} and {\em whole} in an error analysis for named entity recognition.
\begin{figure*}[t]
  \centering
  \includegraphics[width=\textwidth]{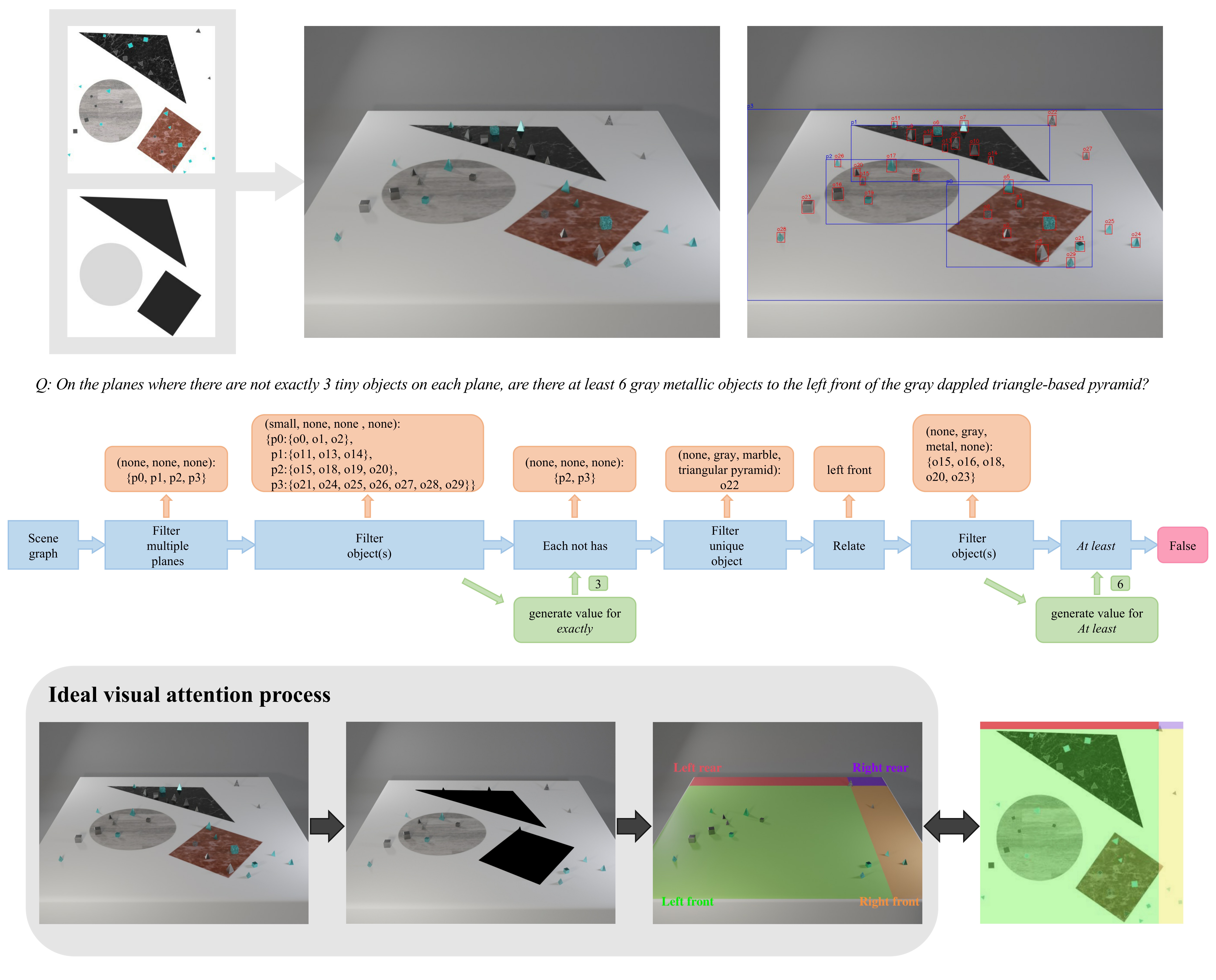}
  \caption{An overview of our dataset. \textbf{Top:} Image generation process and bounding box information. The top two-dimensional image records the scene graph, and the bottom gray-scale color map records roughness of each plane. \textbf{Center:} Examples of questions and their associated operators. \textbf{Bottom:} Ideal visual attention as the operator proceeds.}
  \vspace{-3mm}
  \label{fig:overview}
\end{figure*}

\section{QLEVR}
\label{vqa_dataset}
We design a challenge dataset called QLEVR (for Quantificational Language and Elementary Visual Reasoning) that requires more complex reasoning than previous visual question-answering  datasets. QLEVR is designed to probe the visual reasoning capabilities of visual question-answering systems with respect to quantificational language, including detecting members of sets, quantifying sets, and reasoning about the relationships between sets. To this end, we automatically construct \emph{scene graphs}~\citep{johnson2015image} and use these to generate synthetic images with ground-truth locations, attributes, and relationships for planes and objects. Each scene graph can be queried in a number of way, and we design query templates to render natural language questions involving complex reasoning about sets of such planes and objects. We describe each of these steps in detail: 

\paragraph{Image Generation}
\label{image_generation}
All images in QLEVR are images of objects organized in a particular way on a desk-like surface. Figure~\ref{fig:overview} shows how the images are generated. We construct a scene graph for a two-dimensional image containing areas and objects of different sizes and shapes. Scene graphs determine the ground-truth locations, bounding boxes, attributes and relationships for the planes and objects in the form of a graph or tree structure. Nodes are planes or objects annotated with attributes, each of which is connected to its spatially related nodes. 

Each image contains one to five areas or geometric planes. These can be either triangular, rectangular or circular. The rest of the desk area we refer to as the white non-geometric plane. Geometric planes come in two materials (marble and wood), three colors (black, gray, and brown), and random sizes.

Each geometric plane contains one to ten (1--10) objects, with different sizes and shapes, and the non-geometric plane contains one to twelve (1--12) objects, with different sizes and shapes. Object come in seven shapes (cone, cube, cylinder, pentahedron, sphere, triangular prism, and tetrahedron), two absolute sizes (small and large), five materials (metal, rubber, leather, marble, and wood), and eight colors (blue, brown, cyan, gray, green, purple, red and yellow). The spatial relationships between planes and objects include \emph{front}, \emph{back}, \emph{left} and \emph{right}, as well as  \emph{right front}, \emph{right rear}, \emph{left front} and \emph{left rear}. 

We render three-dimensional images of the scene graphs with Blender~\citep{Blender}. Light settings and three preset camera positions were chosen at random, after validating that all objects were at least partially visible. Since the depth of the scene can affect the judgment of the spatial relationship in the three-dimensional image, the desk boundary is always visible as a reference for determining the depth of the scene. Minimum distances between objects and planes were kept to reduce the ambiguity of spatial relationships. See Appendix~\ref{sec:mat_imgs} for more details. 

\begin{table}[]
    \centering
    \footnotesize 
    \begin{align*}
    &all_P(A, B) \Leftrightarrow A \subseteq B  \\
    &some_P(A, B) \Leftrightarrow A \cap B \neq \emptyset  \\
    &no_P(A, B) \Leftrightarrow A \cap B = \emptyset  \\
    &some\ but\ not\ all_P(A, B) \Leftrightarrow A \cap B \neq \emptyset \neq A - B  \\
    &most_P(A, B) \Leftrightarrow \lvert A \cap B \rvert > \lvert A - B \rvert  \\
    &more_P(A, B) \Leftrightarrow \lvert A \rvert > \lvert B \rvert  \\
    &fewer_P(A, B) \Leftrightarrow \lvert A \rvert < \lvert B \rvert  \\
    &equal_P(A, B) \Leftrightarrow \lvert A \rvert = \lvert B \rvert  \\
    &exactly\ \mathrm{n}_P(A, B) \Leftrightarrow \lvert A \rvert = \mathrm{n} \And A \subseteq B  \\
    &between\ \mathrm{n_1}\ and\ \mathrm{n_2}_P(A, B) \Leftrightarrow \mathrm{n_1} \leq \lvert A \cap B \rvert \leq \mathrm{n_2}  \\
    &at\ most\ \mathrm{n}_P(A, B) \Leftrightarrow \lvert A \cap B \rvert \leq \mathrm{n}  \\
    &more\ than\ \mathrm{n}_P(A, B) \Leftrightarrow \lvert A \cap B \rvert > \mathrm{n}  \\
    &all\ but\ at\ least\ \mathrm{n}_P(A, B) \Leftrightarrow \lvert A - B \rvert \geq \mathrm{n}  \\
    &at\ least\ \frac{\mathrm{n}}{\mathrm{d}}\ of\ the_P(A, B) \Leftrightarrow \frac{\lvert A \cap B \rvert}{\lvert A \rvert}  \geq \frac{\mathrm{n}}{\mathrm{d}}  \\
    &fewer\ than\ \frac{\mathrm{n}}{\mathrm{d}}\ of\ the_P(A, B) \Leftrightarrow \frac{\lvert A \cap B \rvert}{\lvert A \rvert}  < \frac{\mathrm{n}}{\mathrm{d}}  \\
    &no\ objects\ except\ C_P(A, B) \Leftrightarrow A \cap B = \{c\}  \\
    &every\ object\ except\ C_P(A, B) \Leftrightarrow A - B = \{c\} 
\end{align*}
    \caption{Quantifiers included in QLEVR. $P$ denotes the set of all the objects on the target plane(s). $A$ or $B$ denotes a subset of $P$ with the same attributes. $\lvert A \rvert$ is the cardinality of the set $A$.}
    \label{q}
\end{table}

\paragraph{Question Generation}
\label{question_generation}
Quantifiers are often said to be among the most important and complex constructs of natural languages \cite{10.2307/25000951,10.2307/25001052}. 
As pointed out by by \citet{https://doi.org/10.1111/lnc3.12417}, visual question-answering models need to master a wide range of linguistic phenomena, including negation, entailment, mutual exclusivity and so on. We add (generalized) quantifiers to this list and design a dataset to probe the ability of visual question-answering systems to handle quantifiers in combination with other linguistic phenomena. See Table~\ref{q} for the quantifiers included in QLEVR. 

See Figure~\ref{fig:overview} for how questions are formed from scene graphs. In brief, we think of the scene graph as a model and evaluate various combinations of logical operators, including quantifiers, on the scene graph, i.e., performing a model checking \cite{10.1145/1592761.1592781} procedure. 

We introduce the notion of a {\em question family}, defined by a set of  operators and a scene graph. Each question family is associated with 2--6 text templates and a set of synonyms (for shapes, colors, materials, and spatial relationships). The templates were written by hand. Each question template can thus generate multiple questions. For example, the template

\begin{itemize}
    \item[(5)] Are there exactly {\tt <OC> <Z> <C> <M> <S><os>} on the {\tt <PC> <PM> <PS>} plane{\tt <ps>}?
\end{itemize}

\noindent where upper-cased variables refer to words, and lower-cased variables to suffixes, can generate the question

\begin{itemize}
    \item[(6)] Are there exactly 2 small red rubber objects on the black wooden triangular plane?
\end{itemize}

\noindent We construct a total of 671 different templates, which are randomly constructed from 11 plane templates and 61 object templates. Our questions involve attribute recognition, counting, comparing numbers or attributes, spatial relationships, and understanding of quantifiers. Figure~\ref{fig:overview} shows the operators built in the given question family, such as \emph{filter}, \emph{relate}, and \emph{at least}.

Note that many (generalized) quantifiers are related by entailment. 
The question \begin{itemize}
    \item[(7)] Are all the red cubes on the marble planes?
\end{itemize} 
is, assuming an image with red cubes, semantically equivalent to
\begin{itemize}
    \item[(8)] Are no red cubes not on the marble planes?
\end{itemize}

The semantics of combinations of quantifiers can be derived using \emph{squares of opposition}~\citep{Westersthl2012ClassicalVM}. We exploit these entailment relations in creating QLEVR.  

\begin{figure}
    \centering
    \includegraphics[width=3in]{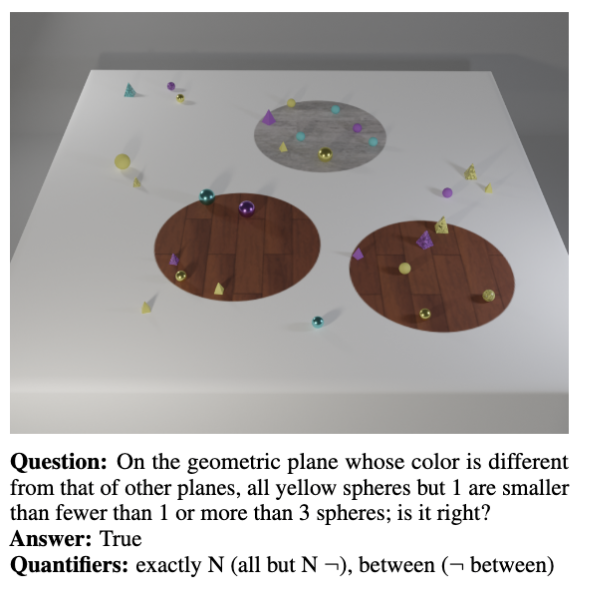}
    \caption{Example image-question pair}
    \label{ex:image}
\end{figure}

Some combinations of key values may generate unreasonable questions. We therefore define restrictions for each question family to avoid the generation of \emph{pragmatically odd}, \emph{ill-posed} or \emph{trivial} questions. For example, the phrase \emph{on the marble plane where there are at least 5 red objects} would be pragmatically odd if there was only one marble plane in the scene. The sentence \begin{itemize}
    \item[(9)] On the marble plane, do between 2 and 4 cubes have the same size as most of the cylinders?
\end{itemize}

\noindent is ill-posed if there are no cubes on the marble plane. Finally, questions like \emph{Are there more red cubes than cubes?} are trivial, because they can be answered in the absence of the image. The assertion is always true. The opposite would, for example, be true of  

\begin{itemize}
    \item[(10)] On the plane with 8 balls, are there exactly 3 balls?
\end{itemize}

We present many examples of images and questions in the Appendix, but see also Figure~\ref{ex:image} for a complex question with embedded quantifiers.

\begin{figure*}[t]
    \centering
    \includegraphics[width=0.41\textwidth]{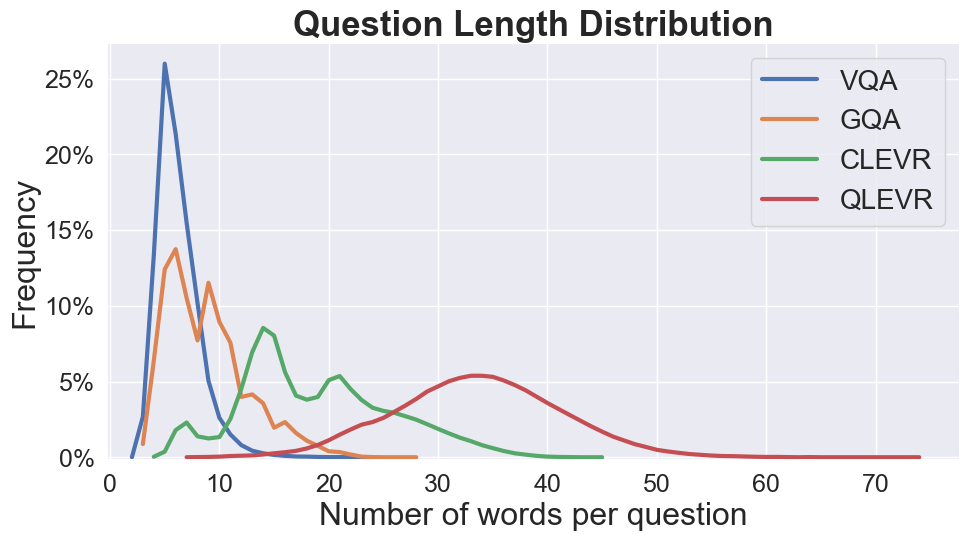} \hspace{2mm}
    \includegraphics[width=0.56\textwidth]{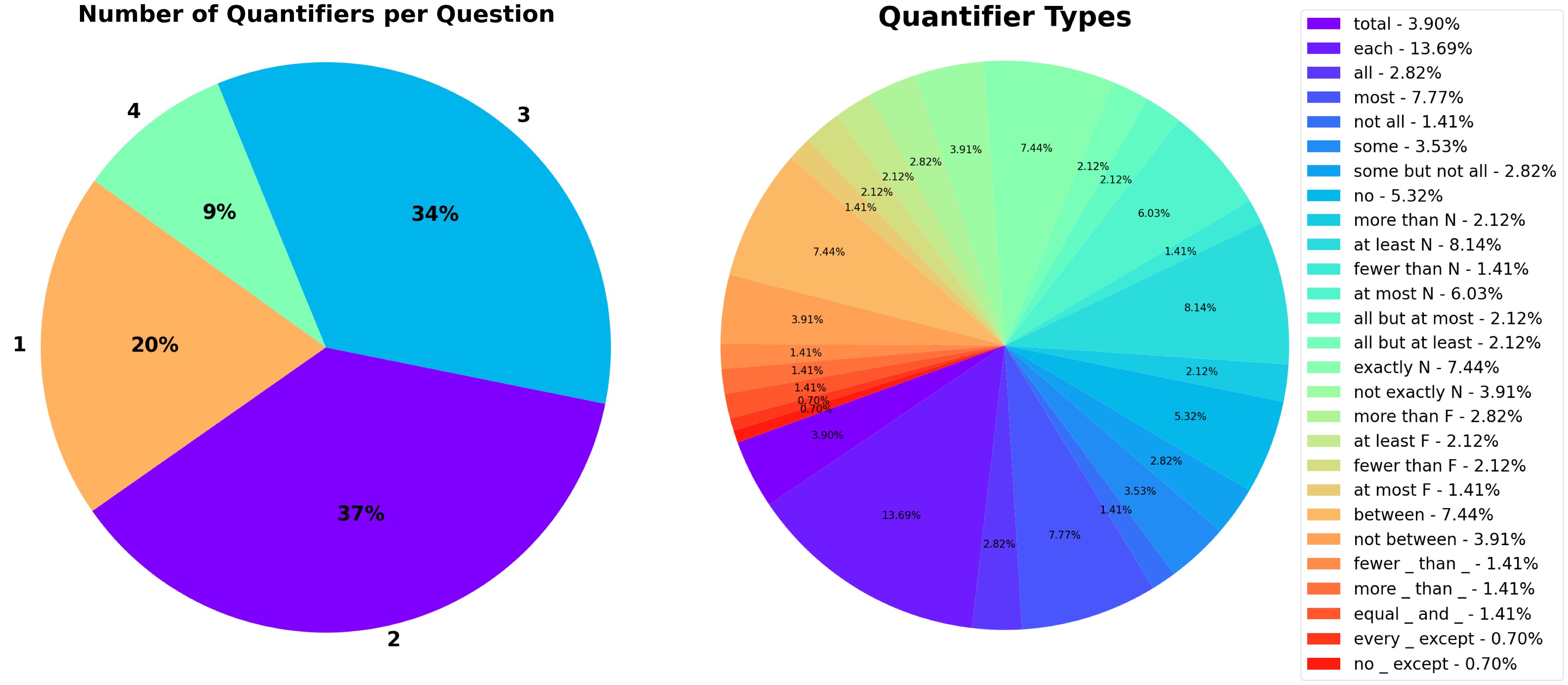}
    \caption{Statistics for our dataset. \textbf{Left:} Question length distribution for different popular VQA datasets; most of the QLEVR questions have 30 to 40 words, which is longer than other datasets. \textbf{Middle:} Distribution of the number of quantifiers in QLEVR questions. \textbf{Right:} Frequency distribution of quantifiers in QLEVR,  where \emph{N} stands for \emph{Number} and \emph{F} stands for \emph{Fraction}; \emph{each}, \emph{total}, \emph{no}, \emph{at most N}, \emph{at least N}, \emph{exactly N}, \emph{between}, \emph{not between}, and \emph{not exactly N} are also used in the plane templates, so they appear more frequently. For the quantifiers in the texts of the same question family, we consider each quantifier in square of opposition \{\emph{Q}, $\emph{Q}\neg$, $\neg\emph{Q}$, $\emph{Q}^\emph{d}$\} as quantifier \emph{Q}.}
    \label{fig:qtf_smry}
\end{figure*}

\begin{table}[t]
  \centering
  \scalebox{0.8}{
    \setlength\tabcolsep{3pt}
    \begin{tabular}{c|ccccc}
		\toprule
        \multirow{2}{*}{Split} 
        &  Images & Questions & \makecell{Unique \\ questions} & \makecell{Overlap \\ with train} & \makecell{Overlap \\ with val} \\
        \hline
        Train    & 70,000 & 700,000    & 699,498      & -  & - \\  
        Val    & 15,000 & 150,000   & 149,968      & 199 | 148 & - \\  
        Test    & 15,000  & 150,000  & 149,980      & 194 | 145  & 49 | 39 \\  
        \hline
        Total    & 100,000  & 1,000,000  & 999,446      & -  & -  \\  
        \bottomrule
	\end{tabular}
  }
\caption{Statistics for our dataset. In each \emph{Overlap} column, the number on the right represents the number of overlapping questions with the same answer.}
  \label{tab:stats}
\end{table}

\paragraph{Dataset Characteristics}
\label{statistics}
QLEVR has 1,000,000 questions for 100,000 images, with each image having 10 questions generated from different question families. The dataset is balanced, preventing answering in the absence of the images. 
In addition, the answer distribution across question families is constrained by acceptance-rejection sampling. 
The data is randomly split, with 70\% for training data, 15\% for validation and 15\% for heldout evaluation data (the test set). As shown in figure~\ref{fig:qtf_smry}, QLEVR includes 27 different quantifiers. Questions contain 1--4 quantifiers. Table~\ref{tab:stats} shows the diversity and complexity of the QLEVR questions. Almost all the questions are unique. Very few questions appear in several splits, and always in conjunction with new scene graphs.

\section{Experiments}
\label{Experiments}
In this section, we evaluate the performance of baselines and near-state-of-the-art models on the QLEVR dataset and perform detailed error analysis. We ran each each method three times with different random seeds and report the test set  performance for the model that achieved the best performance on the validation data.

\subsection{Models}
We first present three purely text-based models, Q-type~\citep{{VQA}}, LSTM~\citep{LSTM} and BERT~\citep{devlin2018bert}, to evaluate the level of visual reasoning needed for QLEVR. If these perform at random (0.5), we have successfully constructed a dataset in which questions cannot be answered in the absence of images. It is important to include text-only models as baselines in visual question answering to control for spurious correlations \cite{NEURIPS2020_20d749bc}. We shall see in \S4.2 that while Q-type performs at chance level, the BERT and LSTM baselines are able to pick up on some spurious correlations. We also evaluate two standard visual question answering architectures, one based on a combination of convolutional and recurrent neural networks (CNN+LSTM), and one attention-based architecture \citep{hudson2018compositional}. The latter performs best on the counting and number comparison tasks in the CLEVR dataset~\citep{johnson2017clevr} compared with other approaches, such as Bottom-Up-Attention and Top-Down (UpDn)~\citep{Anderson2017up-down}, Question-Conditioned Graph (QCG)~\citep{learningconditionedgraph}, Bilinear Attention Network (BAN)~\citep{Kim2018}, Relation Network (RN)~\citep{Santoro2017ASN} and Recurrent Aggregation of Multimodal Embeddings Network (RAMEN) ~\citep{shrestha2019answer}. 
We describe each system in detail: 

\begin{itemize}[leftmargin=*]

\item \textbf{Q-type~\citep{{VQA}}:} Similar to the "per Q-type prior" method in ~\citep{{VQA}}, this baseline predicts the most popular answer for each question type.

\item \textbf{LSTM~\citep{LSTM}:} Question words are embedded as  300-dimensional vector sand fed into an LSTM network. The last hidden state representation is passed into a multi-layer perceptron (MLP) to predict the final answer. All experiments use a bi-directional LSTM with 512 units in the hidden layer per direction. 

\item \textbf{BERT~\citep{devlin2018bert}:} We fine-tune BERT \cite{devlin2018bert} augmented with a sentence-level classification head: The special classification token \texttt{[CLS]} is passed to a feed-forward layer and used for sentence class prediction. 

\item \textbf{CNN+LSTM:} The images are encoded using a convolutional neural network and questions as the last hidden state produced by an LSTM network. The convolutional network uses spatial features produced by ResNet-101~\citep{he2015deep} pre-trained on ImageNet~\citep{5206848}. We resize all images to 448x448, and use the final average pooling layer to extracts features of the shape (1, 14, 14, 2048). The question and image features are concatenated and passed to a multi-layered perceptron to predict the final answer.

\item \textbf{MAC~\citep{hudson2018compositional}:} The MAC network is a recurrent attention network, which uses a Memory, Attention, and Composition (MAC) cell in each attention-based reasoning step to learn to perform iterative reasoning processes. MAC learns compositional reasoning directly from the questions and the images in an end-to-end approach. The word vectors have a dimension of 300 and are initialized randomly using a standard uniform distribution. The images are resized to 448x448, and 2048-dimensional features are produced by ResNet-101. The model uses a hidden state size of 512 and a length of 12 MAC cells.

\end{itemize}

\subsection{Analysis by Quantifier Type}

\begin{table*}[t]
\begin{minipage}{0.48\linewidth}
\centering
\scalebox{.7}{
    \begin{tabular}{lccccc} 
        \toprule
        &\textbf{Q-type} &\textbf{LSTM} &\textbf{BERT} &\textbf{CNN+LSTM} &\textbf{MAC} \\ \midrule 
        each &50.0   &63.9 &65.4 &65.3 &66.2  \\
        total &50.0   &63.4 &64.7 &65.1 &66.3  \\
        all &50.0   &59.5 &60.5 &60.7 &61.3  \\
        most &50.0   &61.7 &63.6 &63.5 &64.0  \\
        not all &50.0   &60.7 &62.2 &61.3 &62.6  \\
        no &50.0   &63.9 &64.6 &64.6 &65.3  \\
        some &50.0   &58.2 &58.9 &58.7 &58.7  \\
        some but not all &50.0   &59.9 &61.5 &61.0 &61.7  \\
        exactly N &50.0   &62.6 &63.8 &63.5 &64.0  \\
        not exactly N &50.0   &64.9 &65.7 &65.9 &66.7  \\
        between &50.0 &65.2   &66.6 &67.0 &67.9  \\
        not between &50.0   &64.4 &65.8 &66.0 &66.0  \\
        all but at most &50.0   &66.9 &68.8 &68.2 &68.9  \\
        all but at least &50.1   &62.7 &63.7 &63.3 &65.2  \\
        \bottomrule
    \end{tabular}
}
\end{minipage}
\hfill
\begin{minipage}{0.48\linewidth}  
\centering
\scalebox{.7}{
    \begin{tabular}{lccccc}        
        \toprule
        &\textbf{Q-type} &\textbf{LSTM} &\textbf{BERT} &\textbf{CNN+LSTM} &\textbf{MAC} \\ \midrule 
        more than N &50.0   &65.5 &67.3 &67.1 &67.2  \\
        at least N &50.0   &64.7 &66.2 &65.8 &66.5  \\
        fewer than N &50.0   &66.7 &67.6 &67.7 &68.3  \\
        at most N &50.0   &64.3 &65.4 &64.9 &65.7  \\
        more than F &50.0   &69.0 &69.6 &71.4 &71.4  \\
        at least F &50.0   &67.9 &70.1 &71.0 &72.0  \\
        fewer than F &50.0   &67.6 &68.8 &70.2 &71.6  \\
        at most F &50.0   &65.8 &68.9 &70.3 &70.8  \\
        every \_ except \_ &50.1   &70.9 &72.0 &70.9 &72.1  \\
        no \_ except \_ &50.1  &78.4 &78.1 &77.9 &78.2  \\
        more \_ than \_ &50.0   &68.0 &69.4 &69.1 &68.7  \\
        fewer \_ than \_ &50.0   &68.9 &69.9 &68.8 &69.0  \\
        equal \_ and \_ &50.0   &61.0 &62.2 &62.0 &62.2  \\
        \textbf{Overall} &\textbf{50.0} &\textbf{64.6} &\textbf{65.8} &\textbf{65.9} & \textbf{66.5}  \\
        \bottomrule
    \end{tabular}
}
\end{minipage}
\caption{Test set results of baselines and state-of-the-art models on the QLEVR dataset. Models are evaluated for both overall accuracy and accuracy per quantifier type. In quantifier type, \emph{N} stands for \emph{Number} and \emph{F} stands for \emph{Fraction}. Refer to Figure~\ref{fig:qtf_smry} for the number distribution of quantifiers.
}
\label{tab:qtf_results}
\end{table*}

Table~\ref{tab:qtf_results} shows the results of the five methods described in \S4.1 on the test set of QLEVR. We make the following observations. 

\begin{itemize}
\item[1.] Q-type exhibits performance levels around 50\% for every quantifier type, showing that the answer distribution of QLEVR is uniform. 
\item[2.] Text-only LSTM and BERT achieve an average accuracies of 64.6\% and 65.8\%, respectively.
These results suggest that even if the answers of each question family are distributed uniformly, there may still be spurious correlations: Objects with more detailed attribute descriptions may be more likely to get a \emph{false} answer. For example, the question \emph{"Are there more than 3 small blue cubes on the black planes?"} is more likely to get a \emph{false} answer than \emph{"Are there more than 3 blue objects on the black planes?"}). 
\item[3.] The CNN+LSTM architecture performs better than LSTM on 24 out of 27 quantifier types and on par with BERT; MAC performs better than LSTM on 26 out of 27 quantifier types,  better than BERT on 24 out of 27 quantifier types and better than CNN+LSTM on 23 out of 27 quantifier types. In general, however, CNN+LSTM and MAC do not improve much over text-only LSTM and BERT, suggesting that the image features extracted by ResNet-101 contain little information relevant to counting in complex scenes.
\item[4.] Accuracies for quantifiers that present thresholds (e.g., \emph{more than N, at least F, etc.}) are higher than for quantifiers that require a number of objects to match exact values (e.g., \emph{exactly N}).
\item[5.] Quantifiers without numerals (e.g., \emph{all, most, not all, some} and \emph{some but not all}) lead to lower accuracies than other quantifiers, showing that reasoning with these quantifiers is harder. This highlights the need for including such quantifiers in challenge datasets to push advancements in visual question answering. 
\end{itemize}

\begin{figure}[t]
    \centering
    \includegraphics[width=0.48\textwidth]{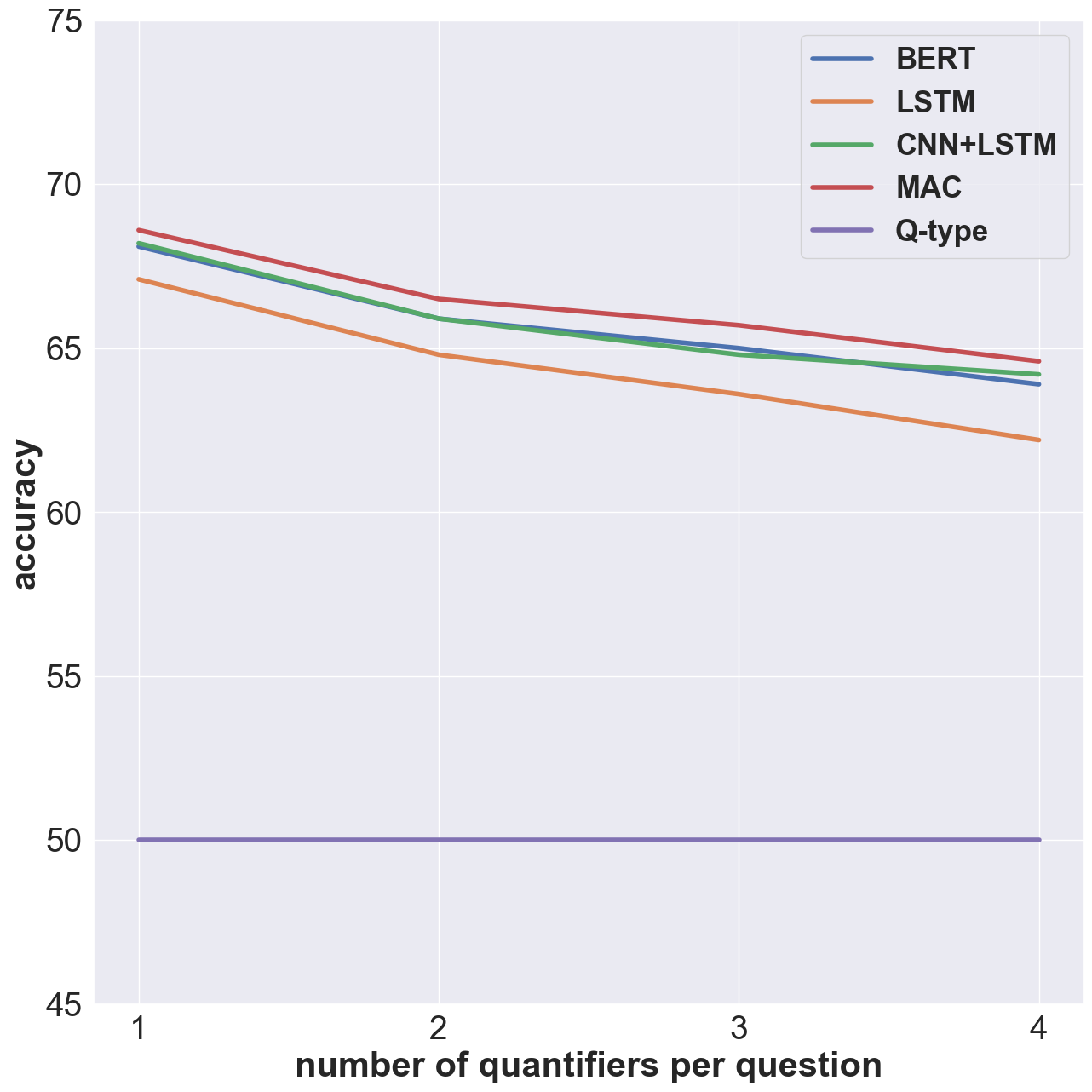}
    \caption{The effect of different number of quantifiers in questions on the accuracy of the answers. Figure~\ref{fig:qtf_smry} shows the distribution of the number
of quantifiers in each QLEVR question.}
    \label{fig:qtf_num_analysis}
\end{figure}

\subsection{Analysis}

\paragraph{Number of Quantifiers in Questions} Figure~\ref{fig:qtf_num_analysis} shows how accuracy varies as the number of quantifiers in the questions increases.
The more quantifiers in a question, the more complex its semantics will be.

\begin{figure*}[t]
    \centering
    \includegraphics[width=0.48\textwidth]{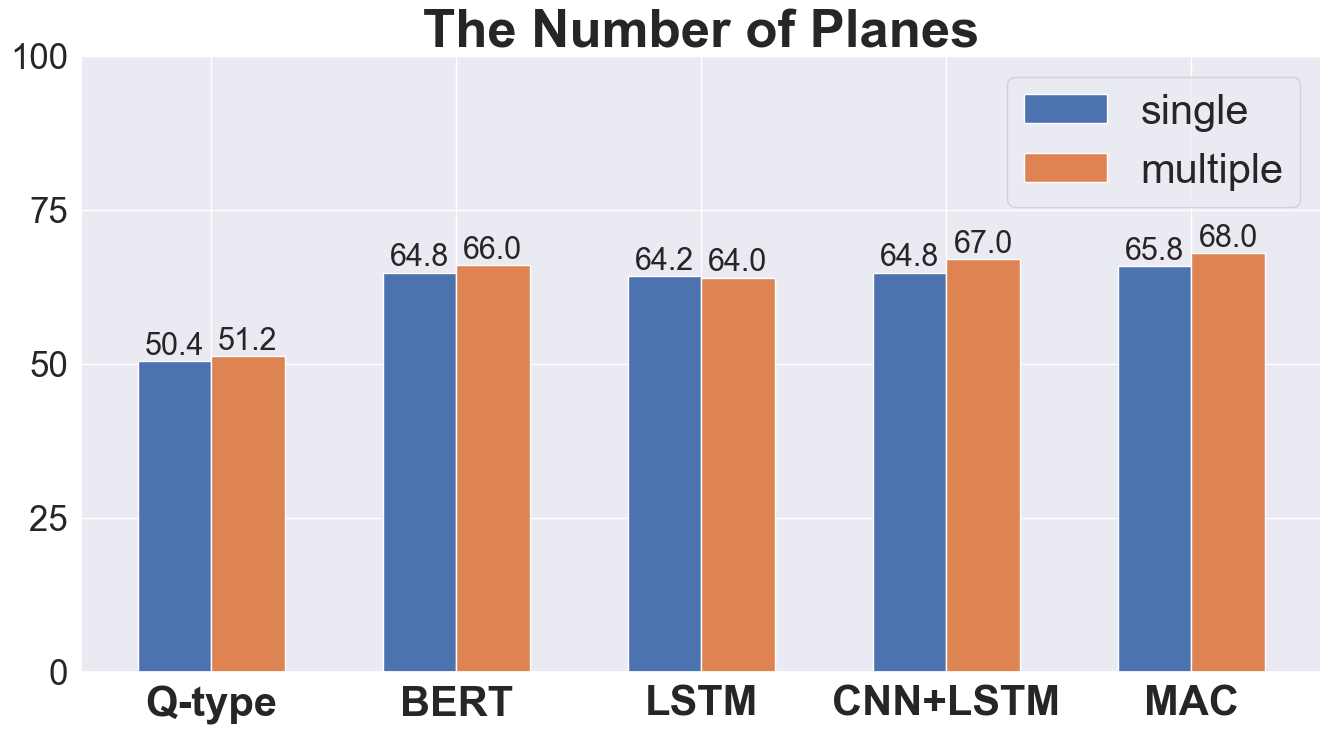} \hspace{2mm}
    \includegraphics[width=0.48\textwidth]{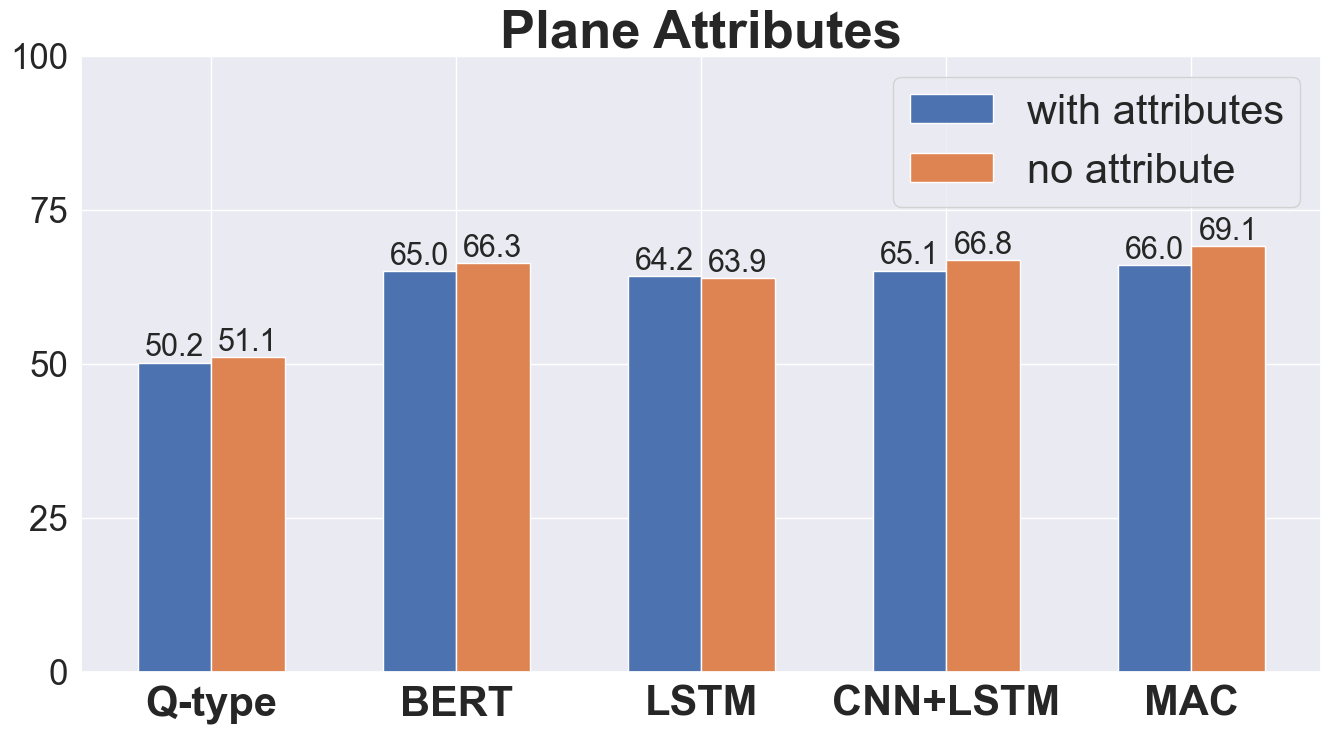}
    \caption{The results of questions contains \emph{"on the <PC> <PM> <PS> plane<ps>"}. \textbf{Left}: The effect of different number of target planes on the accuracy of the answers; \emph{single} means that the reasoning process basically only needs to consider one plane in the image, while \emph{multiple} means that multiple planes need to be considered. \textbf{Right}: The effect of whether the plane has attribute description on the accuracy of the answer; \emph{no attribute ("on the planes")} means that the reasoning process does not need to consider planes in the image and see the image as a whole, while \emph{with attributes ("e.g., on the wooden plane")} means that specific plane(s) needs to be considered.}
    \label{fig:pln_num_analysis}
\end{figure*}

\paragraph{Number of Planes} We also test the visual reasoning abilities of these models by examining error across the number of planes involved in answering the question. Appendix~\ref{sec:q_tpl} introduces all the plane templates in our question families. We use the  plane template \emph{"on the <PC> <PM> <PS> plane<ps>"} for our analysis, because this template has no influence of quantifiers or spatial relationships in targeting planes. QLEVR test set has 13,612 questions with this plane template. The left graph in Figure~\ref{fig:pln_num_analysis} shows how the accuracy varies with the increase in the number of target planes that need to be reasoned with. Among the 13,612 questions, 10,288 of them involve a single plane and 3,324 of them involve multiple planes. We can see that for language-only models Q-type, BERT and LSTM, the number of target planes does not significantly affect the accuracy. However, for CNN+LSTM and MAC, questions involving just a single plane are harder to answer than those involving multiple planes. This is because for visual models, planes enable disambiguation and thereby reduce the required reasoning. The right graph in Figure~\ref{fig:pln_num_analysis} compares accuracy on questions that do not refer to specific planes (\emph{no attribute}), to questions that refer to specific planes (\emph{with attributes}). Among the 13,612 questions, 1,340 questions do not refer to specific planes, whereas 12,272 do. This distinction has little impact on the performance of our text-only models. For CNN+LSTM and MAC, however, examples in the \emph{no attribute} class exhibit higher accuracies than those in \emph{with attributes}. This, again, shows performance is better when less visual reasoning is required. 
\section{Discussion}
\label{Discussion and Future Work}
In this paper, we proposed a dataset, which we call QLEVR -- for Quantificational Language and
Elementary Visual Reasoning. QLEVR probes the ability of visual question-answering systems to reason with quantificational language, including 27 different quantifiers and combinations thereof. It requires complex visual reasoning  to locate the specific planes and understand various relationships between objects. We increase the semantic diversity of the questions by negating quantifiers and by using different templates for semantically equivalent questions. Our analysis highlights how challenging such examples are to visual question-answering systems, and we hope that QLEVR will help guide push research horizons in visual question-answering by zooming in on the challenges posed by quantificational language. 

One fundamental limitation is that QLEVR only considers English questions, and we plan to extent it to other, typologically unrelated languages. Besides, QLEVR can easily be extended by adding new question families, and questions whose answers are not limited to true or false, e.g., with numbers or attributes as answer types. In addition to the three-dimensional images, we also provide two-dimensional images and scene graphs recording the ground-truth information (see Figure~\ref{fig:overview}). It is also possible to generate questions about 2D images by simply modifying our question families. We hope these two datasets can be used for transfer learning for visual question answering in the future.

\bibliography{anthology,custom}

\begin{thebibliography}{35}
\expandafter\ifx\csname natexlab\endcsname\relax\def\natexlab#1{#1}\fi

\bibitem[{Agrawal et~al.(2017)Agrawal, Lu, Antol, Mitchell, Zitnick, Parikh,
  and Batra}]{10.1007/s11263-016-0966-6}
Aishwarya Agrawal, Jiasen Lu, Stanislaw Antol, Margaret Mitchell, C.~Lawrence
  Zitnick, Devi Parikh, and Dhruv Batra. 2017.
\newblock \href {https://doi.org/10.1007/s11263-016-0966-6} {Vqa: Visual
  question answering}.
\newblock \emph{Int. J. Comput. Vision}, 123(1):4–31.

\bibitem[{Anderson et~al.(2018)Anderson, He, Buehler, Teney, Johnson, Gould,
  and Zhang}]{Anderson2017up-down}
Peter Anderson, Xiaodong He, Chris Buehler, Damien Teney, Mark Johnson, Stephen
  Gould, and Lei Zhang. 2018.
\newblock Bottom-up and top-down attention for image captioning and visual
  question answering.
\newblock In \emph{CVPR}.

\bibitem[{Andreas et~al.(2016)Andreas, Rohrbach, Darrell, and Klein}]{7780381}
Jacob Andreas, Marcus Rohrbach, Trevor Darrell, and Dan Klein. 2016.
\newblock \href {https://doi.org/10.1109/CVPR.2016.12} {Neural module
  networks}.
\newblock In \emph{2016 IEEE Conference on Computer Vision and Pattern
  Recognition (CVPR)}, pages 39--48.

\bibitem[{Antol et~al.(2015)Antol, Agrawal, Lu, Mitchell, Batra, Zitnick, and
  Parikh}]{VQA}
Stanislaw Antol, Aishwarya Agrawal, Jiasen Lu, Margaret Mitchell, Dhruv Batra,
  C.~Lawrence Zitnick, and Devi Parikh. 2015.
\newblock {VQA}: {V}isual {Q}uestion {A}nswering.
\newblock In \emph{International Conference on Computer Vision (ICCV)}.

\bibitem[{Barra et~al.(2021)Barra, Bisogni, Marsico, and
  Ricciardi}]{Barra2021VisualQA}
Silvio Barra, Carmen Bisogni, Maria~De Marsico, and Stefano Ricciardi. 2021.
\newblock Visual question answering: which investigated applications?
\newblock \emph{Pattern Recognit. Lett.}, 151:325--331.

\bibitem[{Barwise and Cooper(1981)}]{10.2307/25001052}
Jon Barwise and Robin Cooper. 1981.
\newblock \href {http://www.jstor.org/stable/25001052} {Generalized quantifiers
  and natural language}.
\newblock \emph{Linguistics and Philosophy}, 4(2):159--219.

\bibitem[{Bernardi and Pezzelle(2021)}]{https://doi.org/10.1111/lnc3.12417}
Raffaella Bernardi and Sandro Pezzelle. 2021.
\newblock \href {https://doi.org/https://doi.org/10.1111/lnc3.12417}
  {Linguistic issues behind visual question answering}.
\newblock \emph{Language and Linguistics Compass}, 15(6):e12417.

\bibitem[{Clarke et~al.(2009)Clarke, Emerson, and
  Sifakis}]{10.1145/1592761.1592781}
Edmund~M. Clarke, E.~Allen Emerson, and Joseph Sifakis. 2009.
\newblock \href {https://doi.org/10.1145/1592761.1592781} {Model checking:
  Algorithmic verification and debugging}.
\newblock \emph{Commun. ACM}, 52(11):74–84.

\bibitem[{Community(2018)}]{Blender}
Blender~Online Community. 2018.
\newblock \href {http://www.blender.org} {\emph{Blender - a 3D modelling and
  rendering package}}.
\newblock Blender Foundation, Stichting Blender Foundation, Amsterdam.

\bibitem[{Deng et~al.(2009)Deng, Dong, Socher, Li, Li, and Fei-Fei}]{5206848}
Jia Deng, Wei Dong, Richard Socher, Li-Jia Li, Kai Li, and Li~Fei-Fei. 2009.
\newblock \href {https://doi.org/10.1109/CVPR.2009.5206848} {Imagenet: A
  large-scale hierarchical image database}.
\newblock In \emph{2009 IEEE Conference on Computer Vision and Pattern
  Recognition}, pages 248--255.

\bibitem[{Devlin et~al.(2019)Devlin, Chang, Lee, and
  Toutanova}]{devlin2018bert}
Jacob Devlin, Ming-Wei Chang, Kenton Lee, and Kristina Toutanova. 2019.
\newblock Bert: Pre-training of deep bidirectional transformers for language
  understanding.
\newblock \emph{NAACL-HLT}.

\bibitem[{Dua et~al.(2019)Dua, Wang, Dasigi, Stanovsky, Singh, and
  Gardner}]{dua-etal-2019-drop}
Dheeru Dua, Yizhong Wang, Pradeep Dasigi, Gabriel Stanovsky, Sameer Singh, and
  Matt Gardner. 2019.
\newblock \href {https://doi.org/10.18653/v1/N19-1246} {{DROP}: A reading
  comprehension benchmark requiring discrete reasoning over paragraphs}.
\newblock In \emph{Proceedings of the 2019 Conference of the North {A}merican
  Chapter of the Association for Computational Linguistics: Human Language
  Technologies, Volume 1 (Long and Short Papers)}, pages 2368--2378,
  Minneapolis, Minnesota. Association for Computational Linguistics.

\bibitem[{Fang and Lou(2021)}]{Fang2021PartW}
Lei Fang and Jian-Guang Lou. 2021.
\newblock Part \& whole extraction: Towards a deep understanding of
  quantitative facts for percentages in text.
\newblock \emph{ArXiv}, abs/2110.13505.

\bibitem[{Gat et~al.(2020)Gat, Schwartz, Schwing, and
  Hazan}]{NEURIPS2020_20d749bc}
Itai Gat, Idan Schwartz, Alexander Schwing, and Tamir Hazan. 2020.
\newblock \href
  {https://proceedings.neurips.cc/paper/2020/file/20d749bc05f47d2bd3026ce457dcfd8e-Paper.pdf}
  {Removing bias in multi-modal classifiers: Regularization by maximizing
  functional entropies}.
\newblock In \emph{Advances in Neural Information Processing Systems},
  volume~33, pages 3197--3208. Curran Associates, Inc.

\bibitem[{Hassantabar(2018)}]{Hassantabar2018VisualQA}
Shayan Hassantabar. 2018.
\newblock Visual question answering : Datasets , methods , challenges and
  oppurtunities.

\bibitem[{He et~al.(2015)He, Zhang, Ren, and Sun}]{he2015deep}
Kaiming He, Xiangyu Zhang, Shaoqing Ren, and Jian Sun. 2015.
\newblock \href {http://arxiv.org/abs/1512.03385} {Deep residual learning for
  image recognition}.

\bibitem[{Hintikka(1977)}]{10.2307/25000951}
Jaakko Hintikka. 1977.
\newblock \href {http://www.jstor.org/stable/25000951} {Quantifiers in natural
  languages: Some logical problems ii}.
\newblock \emph{Linguistics and Philosophy}, 1(2):153--172.

\bibitem[{Hochreiter and Schmidhuber(1997)}]{LSTM}
Sepp Hochreiter and J{\"u}rgen Schmidhuber. 1997.
\newblock Long short-term memory.
\newblock \emph{Neural computation}.

\bibitem[{Hudson and Manning(2018)}]{hudson2018compositional}
Drew~A Hudson and Christopher~D Manning. 2018.
\newblock Compositional attention networks for machine reasoning.

\bibitem[{Johnson et~al.(2017)Johnson, Hariharan, van~der Maaten, Fei-Fei,
  Zitnick, and Girshick}]{johnson2017clevr}
Justin Johnson, Bharath Hariharan, Laurens van~der Maaten, Li~Fei-Fei,
  C~Lawrence Zitnick, and Ross Girshick. 2017.
\newblock Clevr: A diagnostic dataset for compositional language and elementary
  visual reasoning.
\newblock In \emph{CVPR}.

\bibitem[{Johnson et~al.(2015)Johnson, Krishna, Stark, Li, Shamma, Bernstein,
  and Fei-Fei}]{johnson2015image}
Justin Johnson, Ranjay Krishna, Michael Stark, Li-Jia Li, David Shamma, Michael
  Bernstein, and Li~Fei-Fei. 2015.
\newblock Image retrieval using scene graphs.
\newblock In \emph{Proceedings of the IEEE conference on computer vision and
  pattern recognition}, pages 3668--3678.

\bibitem[{Joshi et~al.(2020)Joshi, Aditya, Sathe, and
  Choudhury}]{joshi-etal-2020-taxinli}
Pratik Joshi, Somak Aditya, Aalok Sathe, and Monojit Choudhury. 2020.
\newblock \href {https://doi.org/10.18653/v1/2020.conll-1.4} {{T}axi{NLI}:
  Taking a ride up the {NLU} hill}.
\newblock In \emph{Proceedings of the 24th Conference on Computational Natural
  Language Learning}, pages 41--55, Online. Association for Computational
  Linguistics.

\bibitem[{Kim et~al.(2018)Kim, Jun, and Zhang}]{Kim2018}
Jin-Hwa Kim, Jaehyun Jun, and Byoung-Tak Zhang. 2018.
\newblock {Bilinear Attention Networks}.
\newblock In \emph{Advances in Neural Information Processing Systems 31}, pages
  1571--1581.

\bibitem[{Liu et~al.(2021)Liu, Bugliarello, Ponti, Reddy, Collier, and
  Elliott}]{liu-etal-2021-visually}
Fangyu Liu, Emanuele Bugliarello, Edoardo~Maria Ponti, Siva Reddy, Nigel
  Collier, and Desmond Elliott. 2021.
\newblock \href {https://doi.org/10.18653/v1/2021.emnlp-main.818} {Visually
  grounded reasoning across languages and cultures}.
\newblock In \emph{Proceedings of the 2021 Conference on Empirical Methods in
  Natural Language Processing}, pages 10467--10485, Online and Punta Cana,
  Dominican Republic. Association for Computational Linguistics.

\bibitem[{Lu et~al.(2016)Lu, Yang, Batra, and Parikh}]{NIPS2016_9dcb88e0}
Jiasen Lu, Jianwei Yang, Dhruv Batra, and Devi Parikh. 2016.
\newblock \href
  {https://proceedings.neurips.cc/paper/2016/file/9dcb88e0137649590b755372b040afad-Paper.pdf}
  {Hierarchical question-image co-attention for visual question answering}.
\newblock In \emph{Advances in Neural Information Processing Systems},
  volume~29. Curran Associates, Inc.

\bibitem[{Malinowski and Fritz(2014)}]{conf/nips/MalinowskiF14}
Mateusz Malinowski and Mario Fritz. 2014.
\newblock \href
  {http://dblp.uni-trier.de/db/conf/nips/nips2014.html#MalinowskiF14} {A
  multi-world approach to question answering about real-world scenes based on
  uncertain input.}
\newblock In \emph{NeurIPS}, pages 1682--1690.

\bibitem[{Norcliffe-Brown et~al.(2018)Norcliffe-Brown, Vafeias, and
  Parisot}]{learningconditionedgraph}
Will Norcliffe-Brown, Efstathios Vafeias, and Sarah Parisot. 2018.
\newblock Learning conditioned graph structures for interpretable visual
  question answering.
\newblock \emph{arXiv preprint arXiv:1806.07243}.

\bibitem[{Parfenova et~al.(2021)Parfenova, Elliott, Fern{\'a}ndez, and
  Pezzelle}]{parfenova-etal-2021-probing}
Iuliia Parfenova, Desmond Elliott, Raquel Fern{\'a}ndez, and Sandro Pezzelle.
  2021.
\newblock \href {https://doi.org/10.18653/v1/2021.repl4nlp-1.16} {Probing
  cross-modal representations in multi-step relational reasoning}.
\newblock In \emph{Proceedings of the 6th Workshop on Representation Learning
  for NLP (RepL4NLP-2021)}, pages 152--162, Online. Association for
  Computational Linguistics.

\bibitem[{Pezzelle and Fern{\'a}ndez(2019)}]{pezzelle-fernandez-2019-red}
Sandro Pezzelle and Raquel Fern{\'a}ndez. 2019.
\newblock \href {https://doi.org/10.18653/v1/D19-1285} {Is the red square big?
  {MAL}e{V}i{C}: Modeling adjectives leveraging visual contexts}.
\newblock In \emph{Proceedings of the 2019 Conference on Empirical Methods in
  Natural Language Processing and the 9th International Joint Conference on
  Natural Language Processing (EMNLP-IJCNLP)}, pages 2865--2876, Hong Kong,
  China. Association for Computational Linguistics.

\bibitem[{Ramakrishnan et~al.(2018)Ramakrishnan, Agrawal, and
  Lee}]{NEURIPS2018_67d96d45}
Sainandan Ramakrishnan, Aishwarya Agrawal, and Stefan Lee. 2018.
\newblock \href
  {https://proceedings.neurips.cc/paper/2018/file/67d96d458abdef21792e6d8e590244e7-Paper.pdf}
  {Overcoming language priors in visual question answering with adversarial
  regularization}.
\newblock In \emph{Advances in Neural Information Processing Systems},
  volume~31. Curran Associates, Inc.

\bibitem[{Sampat et~al.(2021)Sampat, Kumar, Yang, and
  Baral}]{sampat-etal-2021-clevr}
Shailaja~Keyur Sampat, Akshay Kumar, Yezhou Yang, and Chitta Baral. 2021.
\newblock \href {https://doi.org/10.18653/v1/2021.naacl-main.289}
  {{CLEVR}{\_}{HYP}: A challenge dataset and baselines for visual question
  answering with hypothetical actions over images}.
\newblock In \emph{Proceedings of the 2021 Conference of the North American
  Chapter of the Association for Computational Linguistics: Human Language
  Technologies}, pages 3692--3709, Online. Association for Computational
  Linguistics.

\bibitem[{Santoro et~al.(2017)Santoro, Raposo, Barrett, Malinowski, Pascanu,
  Battaglia, and Lillicrap}]{Santoro2017ASN}
Adam Santoro, David Raposo, David G.~T. Barrett, Mateusz Malinowski, Razvan
  Pascanu, Peter~W. Battaglia, and Timothy~P. Lillicrap. 2017.
\newblock A simple neural network module for relational reasoning.
\newblock In \emph{NIPS}.

\bibitem[{Schwartz et~al.(2017)Schwartz, Schwing, and
  Hazan}]{NIPS2017_05192834}
Idan Schwartz, Alexander Schwing, and Tamir Hazan. 2017.
\newblock \href
  {https://proceedings.neurips.cc/paper/2017/file/051928341be67dcba03f0e04104d9047-Paper.pdf}
  {High-order attention models for visual question answering}.
\newblock In \emph{Advances in Neural Information Processing Systems},
  volume~30. Curran Associates, Inc.

\bibitem[{Shrestha et~al.(2019)Shrestha, Kafle, and Kanan}]{shrestha2019answer}
Robik Shrestha, Kushal Kafle, and Christopher Kanan. 2019.
\newblock Answer them all! toward universal visual question answering models.
\newblock In \emph{CVPR}.

\bibitem[{Westerst{\aa}hl(2012)}]{Westersthl2012ClassicalVM}
Dag Westerst{\aa}hl. 2012.
\newblock Classical vs. modern squares of opposition, and beyond.

\end{thebibliography}
\bibliographystyle{acl_natbib}

\clearpage
\pagebreak
\appendix
\begin{figure}[H]
 \centering
 \Large\textbf{Supplementary Material}
\end{figure}
\label{sec:appendix}
\section{Question Templates}
\label{sec:q_tpl}
As described in Section~\ref{question_generation}, QLEVR question templates are composed of 11 plane templates and 61 object templates randomly paired. In this section we detail the difference between these templates.

\paragraph{Plane Templates.} 
The role of the plane templates is to raise our question for specific planes (regions) in the image through some restrictions (\textcolor{cyan}{attributes}, \textcolor{orange}{spatial relations} and \textcolor{purple}{explicitly restricted quantifier phrases}). Basically, the plane templates can generate questions with following types:

\begin{packed_item}
    \item On the \textcolor{cyan}{white non-geometric} planes.
    \item On the \textcolor{cyan}{geometric} plane with a \textcolor{cyan}{different shape (color/material)} from other planes.
    \item On the \textcolor{cyan}{black} planes \textcolor{orange}{to the left rear of the circular plane}.
    \item On the planes where there are \textcolor{purple}{at least 3 red cubes} on \textcolor{purple}{each plane}.
    \item On the \textcolor{cyan}{quadrilateral} plane where there are \textcolor{purple}{at most 5 blue balls}.
    \item On the \textcolor{cyan}{brown} planes where there are \textcolor{purple}{between 1 and 4 triangular prisms} on \textcolor{purple}{each plane}.
    \item On the \textcolor{cyan}{triangular} plane where there is \textcolor{purple}{exactly 1 leathery object}.
    \item On the plane where there are \textcolor{purple}{not 2 to 4 triangular prisms}.
    \item On the \textcolor{cyan}{gray} planes where there are \textcolor{purple}{not exactly 3 items} on \textcolor{purple}{each plane}.
    \item On the \textcolor{cyan}{marble} plane where there are \textcolor{purple}{not any wooden cones}.
    \item On the \textcolor{cyan}{wooden} plane where there is \textcolor{purple}{a total of 7 small rubber objects}.
\end{packed_item}

To avoid pragmatically odd questions, we ensure that the number of planes obtained by the plane templates with restrictions of \textcolor{orange}{spatial relations} and \textcolor{purple}{explicitly restricted quantifier phrases} (e.g. \emph{On the brown planes behind the gray plane}, or \emph{On the brown plane where there are exactly 3 balls}) is less than the number of planes obtained by the templates without these restrictions (e.g. \emph{On the brown planes}) for the same scene graph.

\paragraph{Object Templates.} 
We can use the \emph{operators} representation of the questions templates to analyze model performance on the following forms of reasoning:

\begin{itemize}
    \item \textbf{Existence type 1:} Questions ask whether a certain type of quantifier-restricted object exists on one or some specific planes (\emph{e.g., "Whether all the cyan cubes} [Plane Template]\emph{?"}).
    
    \item \textbf{Existence type 2:} Questions ask whether a certain type of quantifier-restricted object exists in a certain direction of a unique object (\emph{e.g., "}[Plane Template]\emph{, are there fewer than 3 balls behind the cyan cube?"}).
    
    \item \textbf{Comparing attributes:} Questions ask whether two types of quantifier-restricted objects have the same value for some attributes (\emph{e.g., "}[Plane Template]\emph{, is there any small cylinders that has the same color as most leathery tetrahedrons?"}).
    
    \item \textbf{Quantity comparison:} Questions compare the size of two sets of objects (\emph{e.g., "}[Plane Template]\emph{, are there more big blocks than rubber balls?"}).
    
    \item \textbf{Size comparison:} Questions ask which of two quantifier-restricted objects has a larger size (\emph{e.g., "}[Plane Template]\emph{, some red cones are larger than some but not all of the metal cones; is it right?"}).
    
    \item \textbf{Spatial relations:} Questions involves the spatial relationship between objects (\emph{e.g., "}[Plane Template]\emph{, are there more big blocks in front of the yellow cylinder than rubber balls to the left rear of the small block?"}).
\end{itemize}

\begin{figure}[t]
    \centering
    \includegraphics[width=0.48\textwidth]{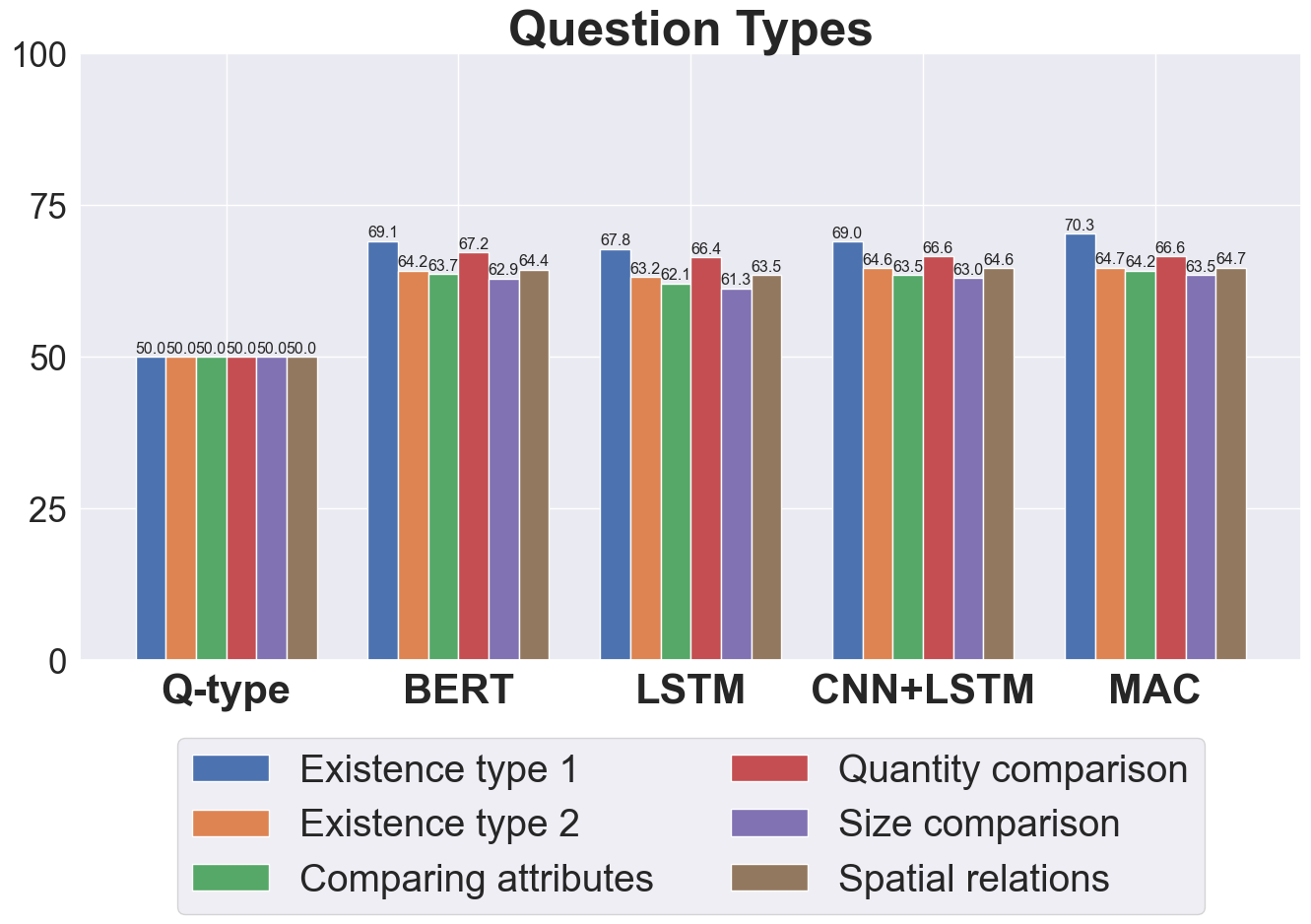}
    \caption{Accuracy per question type on the QLEVR dataset.}
    \label{fig:question_types_analysis}
\end{figure}

Figure~\ref{fig:question_types_analysis} shows the performance on above question types. As can be seen, MAC outperforms other models on most question types. The only exception is: on quantity comparison task, BERT performs slightly better than MAC, showing that MAC has better reasoning ability in complex scenes. Questions of \emph{Existence type 1} obtain better results than \emph{Existence type 2} for vision-language model CNN+LSTM and MAC, suggesting that the position relationship between object and plane is easier to be inferred by the models than the spatial relationship between the objects. For questions of \emph{Quantity comparison}, MAC and CNN+LSTM performs on par with LSTM, suggesting that the image features extracted by ResNet-101 may contain little information related to counting in complex scenes.

\section{3D Modeling and Design}
\label{sec:mat_imgs}
\begin{figure}[t]
    \centering
    \subfigcapskip=-2pt
    \subfigure[object shapes]{
        \begin{minipage}[t]{\linewidth}
        \centering
        \includegraphics[width=0.13\linewidth]{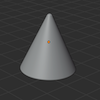}
        \includegraphics[width=0.13\linewidth]{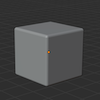}
        \includegraphics[width=0.13\linewidth]{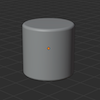}
        \includegraphics[width=0.13\linewidth]{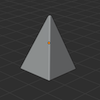}
        \includegraphics[width=0.13\linewidth]{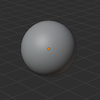}
        \includegraphics[width=0.13\linewidth]{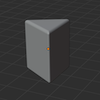}
        \includegraphics[width=0.13\linewidth]{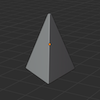}
        \end{minipage}
    }
    
    \subfigure[materials and colors of the planes]{
        \begin{minipage}[t]{\linewidth}
        \centering
        \includegraphics[width=0.15\linewidth]{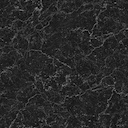}
        \includegraphics[width=0.15\linewidth]{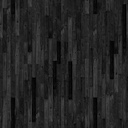}
        \includegraphics[width=0.15\linewidth]{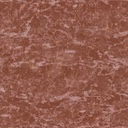}
        \includegraphics[width=0.15\linewidth]{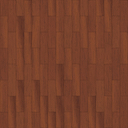}
        \includegraphics[width=0.15\linewidth]{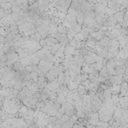}
        \includegraphics[width=0.15\linewidth]{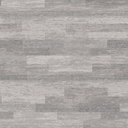}
        \end{minipage}
    }
    
     \subfigure[metal]{
        \begin{minipage}[t]{\linewidth}
        \centering
        \includegraphics[width=0.11\linewidth]{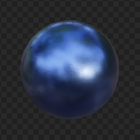}
        \includegraphics[width=0.11\linewidth]{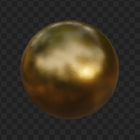}
        \includegraphics[width=0.11\linewidth]{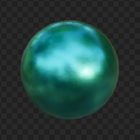}
        \includegraphics[width=0.11\linewidth]{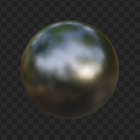}
        \includegraphics[width=0.11\linewidth]{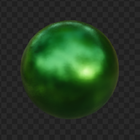}
        \includegraphics[width=0.11\linewidth]{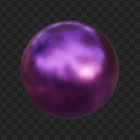}
        \includegraphics[width=0.11\linewidth]{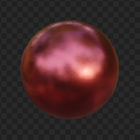}
        \includegraphics[width=0.11\linewidth]{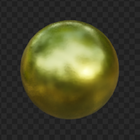}
        \end{minipage}
    }
    
    \subfigure[rubber]{
        \begin{minipage}[t]{\linewidth}
        \centering
        \includegraphics[width=0.11\linewidth]{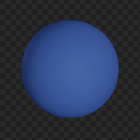}
        \includegraphics[width=0.11\linewidth]{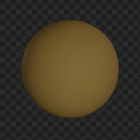}
        \includegraphics[width=0.11\linewidth]{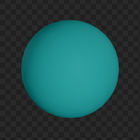}
        \includegraphics[width=0.11\linewidth]{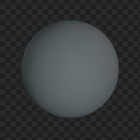}
        \includegraphics[width=0.11\linewidth]{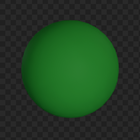}
        \includegraphics[width=0.11\linewidth]{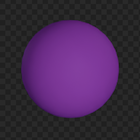}
        \includegraphics[width=0.11\linewidth]{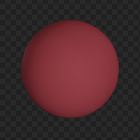}
        \includegraphics[width=0.11\linewidth]{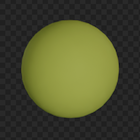}
        \end{minipage}
    }
    
    \subfigure[leather]{
        \begin{minipage}[t]{\linewidth}
        \centering
        \includegraphics[width=0.11\linewidth]{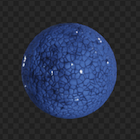}
        \includegraphics[width=0.11\linewidth]{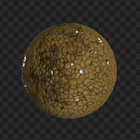}
        \includegraphics[width=0.11\linewidth]{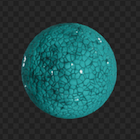}
        \includegraphics[width=0.11\linewidth]{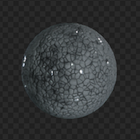}
        \includegraphics[width=0.11\linewidth]{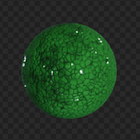}
        \includegraphics[width=0.11\linewidth]{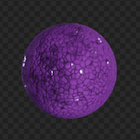}
        \includegraphics[width=0.11\linewidth]{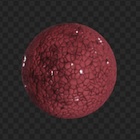}
        \includegraphics[width=0.11\linewidth]{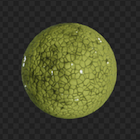} \\ \vspace{0.6mm}
        
        \includegraphics[width=0.11\linewidth]{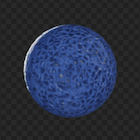}
        \includegraphics[width=0.11\linewidth]{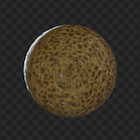}
        \includegraphics[width=0.11\linewidth]{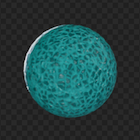}
        \includegraphics[width=0.11\linewidth]{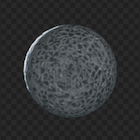}
        \includegraphics[width=0.11\linewidth]{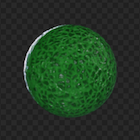}
        \includegraphics[width=0.11\linewidth]{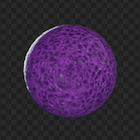}
        \includegraphics[width=0.11\linewidth]{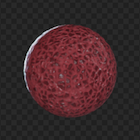}
        \includegraphics[width=0.11\linewidth]{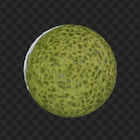} \\ \vspace{2.3mm}
        \end{minipage}
    }
    
    \subfigure[marble]{
        \begin{minipage}[t]{\linewidth}
        \centering
         \includegraphics[width=0.11\linewidth]{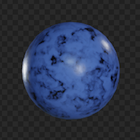}
        \includegraphics[width=0.11\linewidth]{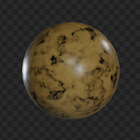}
        \includegraphics[width=0.11\linewidth]{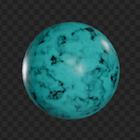}
        \includegraphics[width=0.11\linewidth]{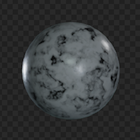}
        \includegraphics[width=0.11\linewidth]{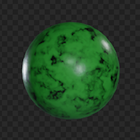}
        \includegraphics[width=0.11\linewidth]{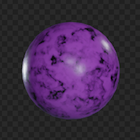}
        \includegraphics[width=0.11\linewidth]{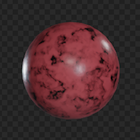}
        \includegraphics[width=0.11\linewidth]{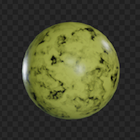} \\ \vspace{0.6mm}
        
        \includegraphics[width=0.11\linewidth]{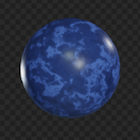}
        \includegraphics[width=0.11\linewidth]{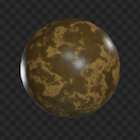}
        \includegraphics[width=0.11\linewidth]{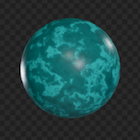}
        \includegraphics[width=0.11\linewidth]{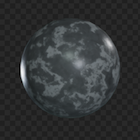}
        \includegraphics[width=0.11\linewidth]{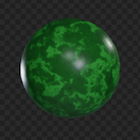}
        \includegraphics[width=0.11\linewidth]{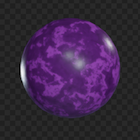}
        \includegraphics[width=0.11\linewidth]{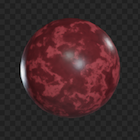}
        \includegraphics[width=0.11\linewidth]{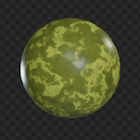}\\ \vspace{2.3mm}
        \end{minipage}
    }
    
    \subfigure[wood]{
        \begin{minipage}[t]{\linewidth}
        \centering
        \includegraphics[width=0.11\linewidth]{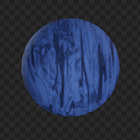}
        \includegraphics[width=0.11\linewidth]{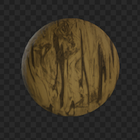}
        \includegraphics[width=0.11\linewidth]{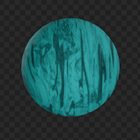}
        \includegraphics[width=0.11\linewidth]{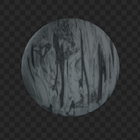}
        \includegraphics[width=0.11\linewidth]{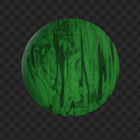}
        \includegraphics[width=0.11\linewidth]{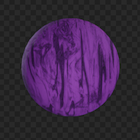}
        \includegraphics[width=0.11\linewidth]{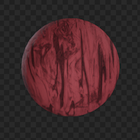}
        \includegraphics[width=0.11\linewidth]{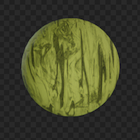} \\ \vspace{0.6mm}
        
        \includegraphics[width=0.11\linewidth]{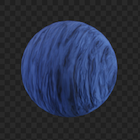}
        \includegraphics[width=0.11\linewidth]{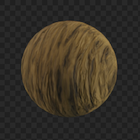}
        \includegraphics[width=0.11\linewidth]{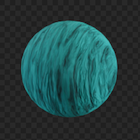}
        \includegraphics[width=0.11\linewidth]{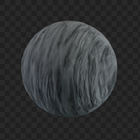}
        \includegraphics[width=0.11\linewidth]{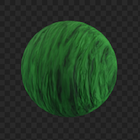}
        \includegraphics[width=0.11\linewidth]{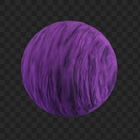}
        \includegraphics[width=0.11\linewidth]{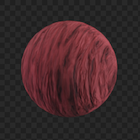}
        \includegraphics[width=0.11\linewidth]{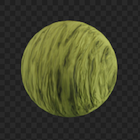}\\ \vspace{2.3mm}
        \end{minipage}
    }
    
    \caption{From left to right, the object shapes in (\textbf{a}) are cone, cube, cylinder, pentahedron, sphere, triangular prism, and tetrahedron; the plane attributes in (\textbf{b}) are black marble, black wood, brown marble, brown wood, gray marble and gray wood; the colors in (\textbf{c}) \textasciitilde (\textbf{g}) are blue, brown, cyan, gray, green, purple, red and yellow.}
    \label{fig:shape_material_color}
\end{figure}

Figure~\ref{fig:shape_material_color} shows the materials and object models made through Blender~\citep{Blender}, as well as the performance of different colors on these materials. Two different materials of leather, marble, and wood were made respectively to further enrich the diversity of objects in the dataset. The images of the plane materials were made by modifying the images under CC0 1.0 Universal.\footnote{\href{https://creativecommons.org/publicdomain/zero/1.0/}{Creative Commons - CC0 1.0 Universal}} Note that after the overall scene rendering, objects of certain materials will produce different effects according to the color and material of the plane in contact with them, as well as the position of the camera and lights.

\section{Example Images and Questions}
\label{sec:example_imgs}
The remaining pages show some images and questions generated by the combination of our different plane templates and object templates. Each question is annotated with its answer and contained quantifiers, where \emph{N} stands for \emph{Number}, \emph{F} stands for \emph{Fraction} and \emph{O} stands for \emph{Object}.
\begin{figure*}
\begin{minipage}{0.48\textwidth}
    \includegraphics[width=\textwidth]{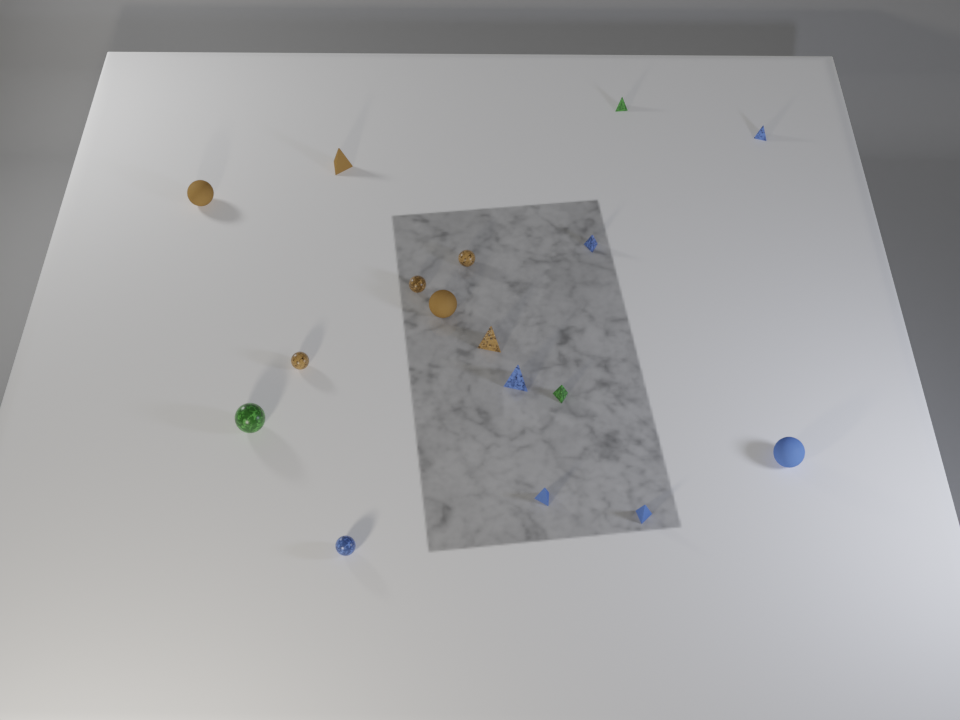}
    \begin{minipage}[t][2.2cm][t]{1\textwidth}
      \footnotesize
      \textbf{Question:} Whether all the large brown objects are on the white plane? \\*
      \textbf{Answer:} False \\*
      \textbf{Quantifiers:} all  \\*[6pt]
    \end{minipage}\\*
    \begin{minipage}[t][2.2cm][t]{1\textwidth}
      \footnotesize
      \textbf{Question:} Some large rubber tetrahedron is not on the gray marble plane; is it right? \\*
      \textbf{Answer:} True \\*
      \textbf{Quantifiers:} not all (some $\neg$)  \\*[6pt]
    \end{minipage}\\*
    \begin{minipage}[t][2.2cm][t]{1\textwidth}
      \footnotesize
      \textbf{Question:} It is not the case that all the big blue rubbery spheres are not on the gray rectangular plane; is it right? \\*
      \textbf{Answer:} False \\*
      \textbf{Quantifiers:} some ($\neg$ all $\neg$) \\*[6pt]
    \end{minipage}
  \end{minipage}
  \hspace{3.5mm}
  \begin{minipage}{0.48\textwidth}
    \includegraphics[width=\textwidth]{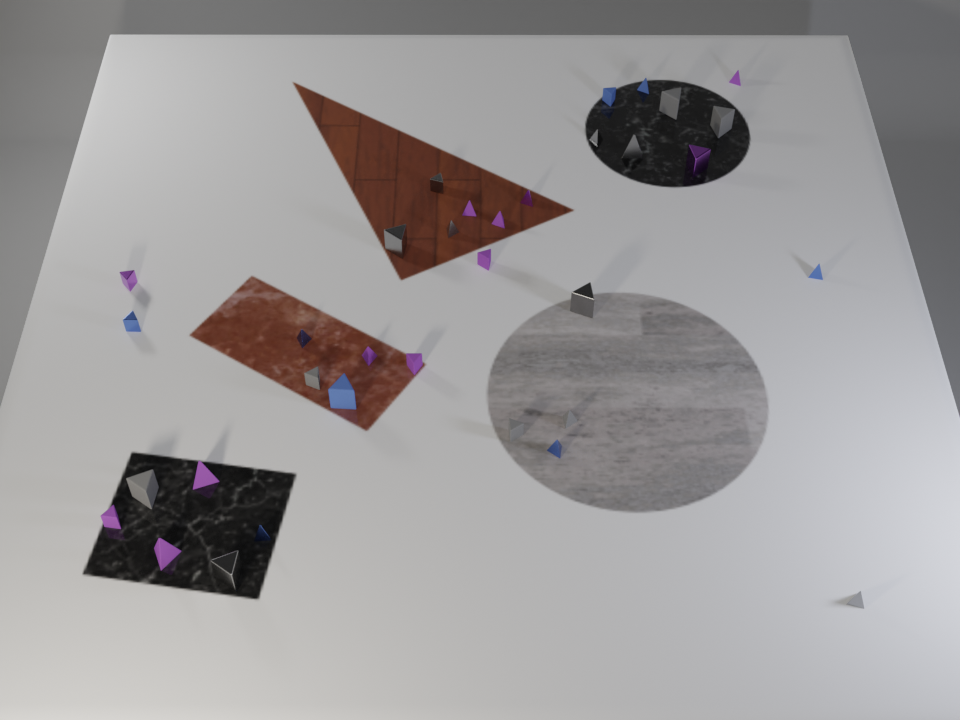}
    \begin{minipage}[t][2.2cm][t]{1\textwidth}
      \footnotesize
      \textbf{Question:} It's not the case that some large purple metallic triangular prism is on the planes where there are 9 items in total; is it right? \\*
      \textbf{Answer:} True \\*
      \textbf{Quantifiers:} total, no ($\neg$ some)  \\*[6pt]
    \end{minipage}\\*
    \begin{minipage}[t][2.2cm][t]{1\textwidth}
      \footnotesize
      \textbf{Question:} Whether some but not all of the large purple rubber objects are on the marble planes where there are 4 blue objects in total? \\*
      \textbf{Answer:} False \\*
      \textbf{Quantifiers:} total, some but not all  \\*[6pt]
    \end{minipage}\\*
    \begin{minipage}[t][2.2cm][t]{1\textwidth}
      \footnotesize
      \textbf{Question:} Are there at most 3 small blue objects on the dappled planes where there are 5 big triangular prisms in total? \\*
      \textbf{Answer:} True \\*
      \textbf{Quantifiers:} total, at most N \\*[6pt]
    \end{minipage}
  \end{minipage}
  \vspace{0.2cm}
  
  \begin{minipage}{0.48\textwidth}
    \includegraphics[width=\textwidth]{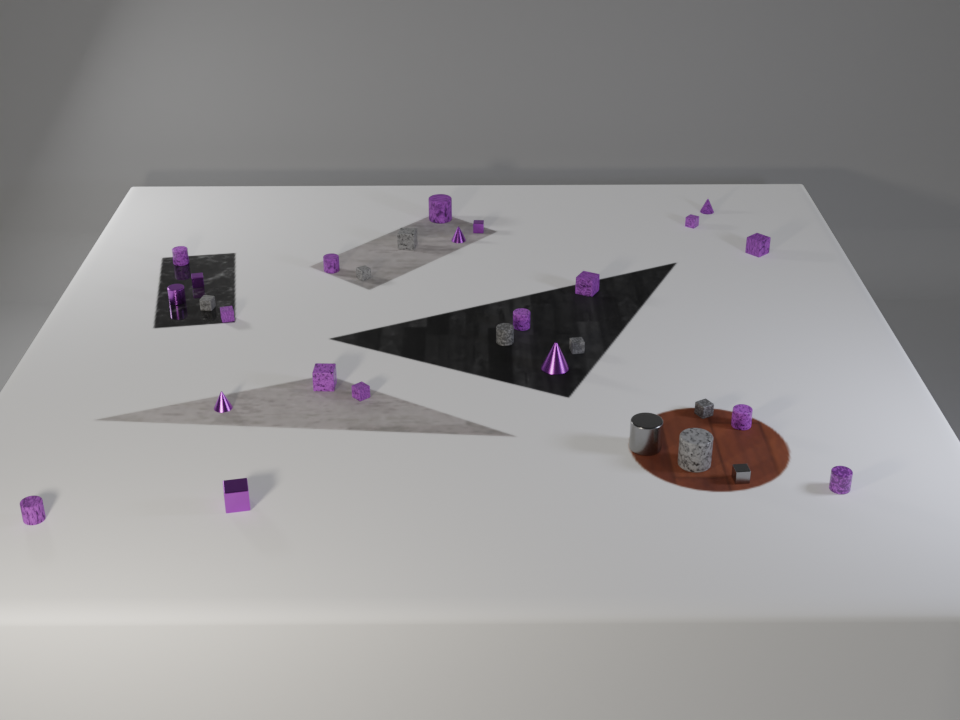}
    \begin{minipage}[t][2.2cm][t]{1\textwidth}
      \footnotesize
      \textbf{Question:} All the big wooden blocks but at least 2 are not on the planes where there are exactly 2 cylinders on each plane; is it right? \\*
      \textbf{Answer:}  True \\*
      \textbf{Quantifiers:} each, exactly N, at least N (all but at least N $\neg$) \\*[6pt]
    \end{minipage}\\*
    \begin{minipage}[t][2.2cm][t]{1\textwidth}
      \footnotesize
      \textbf{Question:} It is not the case that at most 2 wood cylinders are on the quadrilateral plane where there are exactly 2 purple wood cylinders; is it right? \\*
      \textbf{Answer:} False \\*
      \textbf{Quantifiers:} each, exactly N, more than N ($\neg$ at most N)  \\*[6pt]
    \end{minipage}\\*
    \begin{minipage}[t][2.2cm][t]{1\textwidth}
      \footnotesize
      \textbf{Question:} Are there fewer than 2 small purple dappled cylinders on the planes where there is exactly 1 purple block on each plane? \\*
      \textbf{Answer:} True \\*
      \textbf{Quantifiers:} each, exactly N, fewer than N \\*[6pt]
    \end{minipage}
  \end{minipage}
  \hspace{3.5mm}
  \begin{minipage}{0.48\textwidth}
    \includegraphics[width=\textwidth]{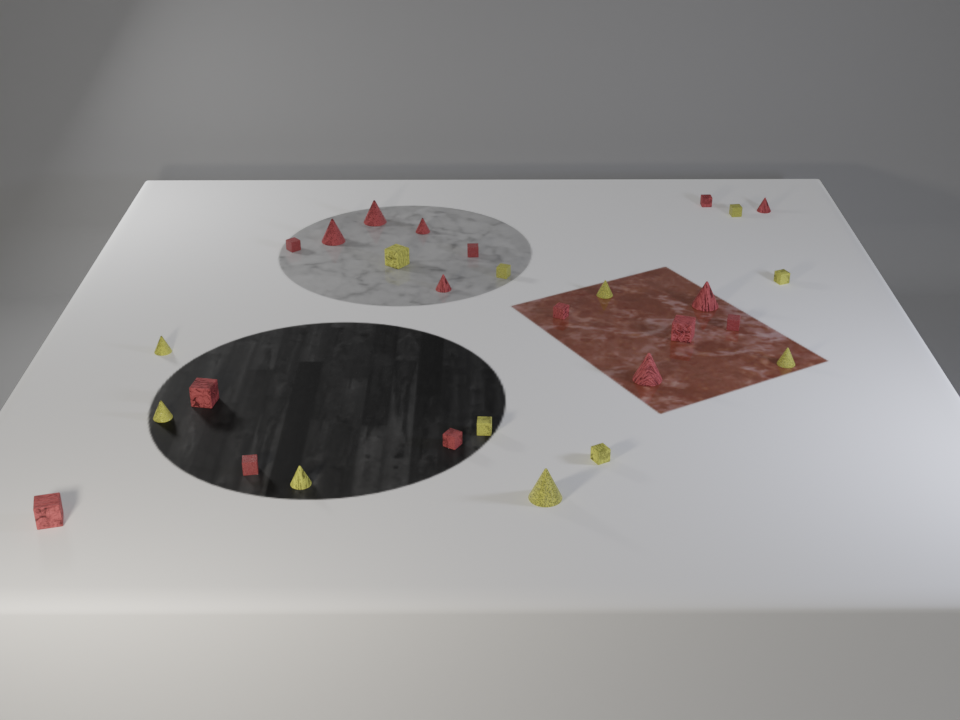}
    \begin{minipage}[t][2.2cm][t]{1\textwidth}
      \footnotesize
      \textbf{Question:} All the tiny red wood objects but 1 are not on the non-white plane where the shape of the plane is different from that of other planes; is it right? \\*
      \textbf{Answer:} True \\*
      \textbf{Quantifiers:} exactly N (all but N $\neg$) \\*[6pt]
    \end{minipage}\\*
    \begin{minipage}[t][2.2cm][t]{1\textwidth}
      \footnotesize
      \textbf{Question:} Are there between 1 and 3 small red leathery cubes on the circular plane to the left rear of the brown quadrilateral plane? \\*
      \textbf{Answer:} True \\*
      \textbf{Quantifiers:} between \\*[6pt]
    \end{minipage}\\*
    \begin{minipage}[t][2.2cm][t]{1\textwidth}
      \footnotesize
      \textbf{Question:} All the red leather objects but at most 3 are on the geometric plane where the material of the plane is different from that of other planes; is it right? \\*
      \textbf{Answer:} False \\*
      \textbf{Quantifiers:} all but at most N \\*[6pt]
    \end{minipage}
  \end{minipage}
\end{figure*}

\begin{figure*}  
\begin{minipage}{0.48\textwidth}
    \includegraphics[width=\textwidth]{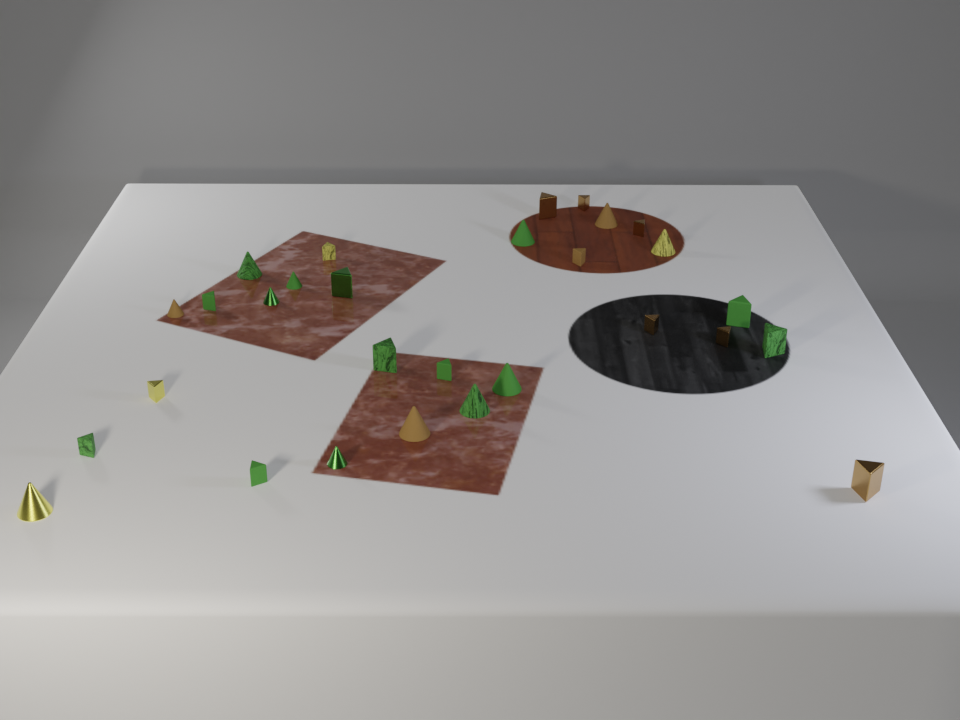}
    \begin{minipage}[t][2.2cm][t]{1\textwidth}
      \footnotesize
      \textbf{Question:} More than two thirds of the green objects are on the brown marble plane to the left front of the black wood plane; is it right? \\*
      \textbf{Answer:} False \\*
      \textbf{Quantifiers:} more than F \\*[6pt]
    \end{minipage}\\*
    \begin{minipage}[t][2.2cm][t]{1\textwidth}
      \footnotesize
      \textbf{Question:} Are most brown metallic objects on the geometric plane on the right side of the brown wooden round plane? \\*
      \textbf{Answer:} False \\*
      \textbf{Quantifiers:} most  \\*[6pt]
    \end{minipage}\\*
    \begin{minipage}[t][2.2cm][t]{1\textwidth}
      \footnotesize
      \textbf{Question:} It is not the case that fewer than 4 brown metallic triangular prisms are not on the non-white plane where the color of the plane is different from that of other planes; is it right? \\*
      \textbf{Answer:} True \\*
      \textbf{Quantifiers:} all but at least N ($\neg$ fewer than N $\neg$) \\*[6pt]
    \end{minipage}
  \end{minipage}
  \hspace{3.5mm}
  \begin{minipage}{0.48\textwidth}
    \includegraphics[width=\textwidth]{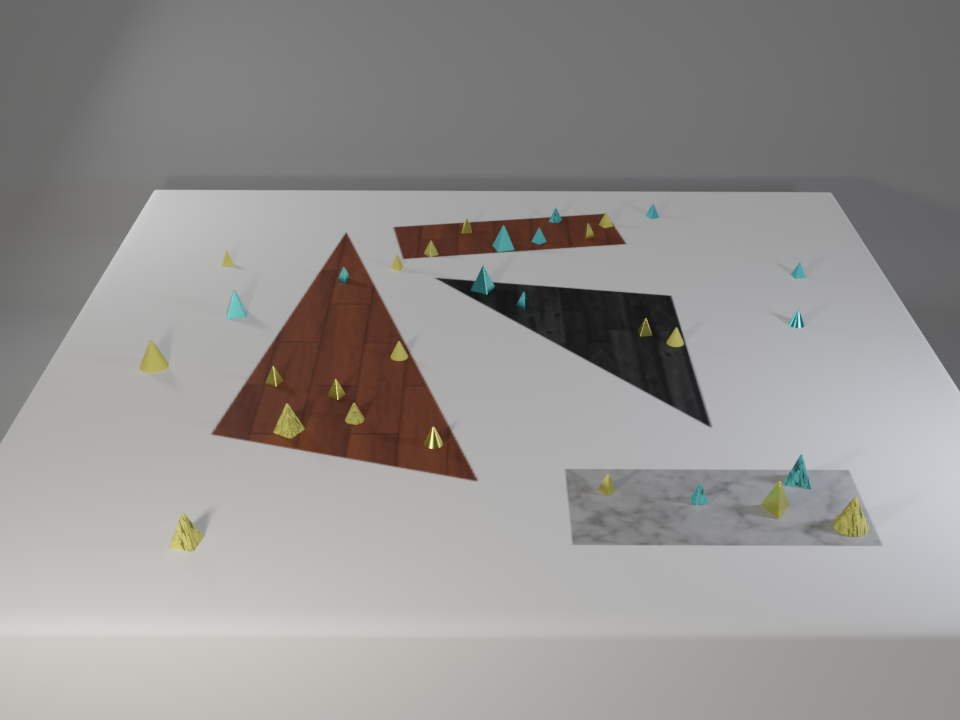}
    \begin{minipage}[t][2.2cm][t]{1\textwidth}
      \footnotesize
      \textbf{Question:} Fewer than three-quarters of the small yellow objects are on the wood three-sided plane where there are not any large metallic square-based pyramids; is it right? \\*
      \textbf{Answer:} True \\*
      \textbf{Quantifiers:} no ($\neg$ any), fewer than F  \\*[6pt]
    \end{minipage}\\*
    \begin{minipage}[t][2.2cm][t]{1\textwidth}
      \footnotesize
      \textbf{Question:} It is not the case that fewer than 11/15 of the metallic items are on the planes where there are 0 tiny rubbery tetrahedrons on each plane; is it right? \\*
      \textbf{Answer:} True \\*
      \textbf{Quantifiers:} each, no (0), at least F ($\neg$ fewer than F) \\*[6pt]
    \end{minipage}\\*
    \begin{minipage}[t][2.2cm][t]{1\textwidth}
      \footnotesize
      \textbf{Question:} At most 7/8 of the yellow rubber triangular pyramids are on the planes where there is no big yellow wooden cone on each plane; is it right? \\*
      \textbf{Answer:} False \\*
      \textbf{Quantifiers:} each, no, at most F \\*[6pt]
    \end{minipage}
  \end{minipage}
  \vspace{0.2cm}
  
  \begin{minipage}{0.48\textwidth}
    \includegraphics[width=\textwidth]{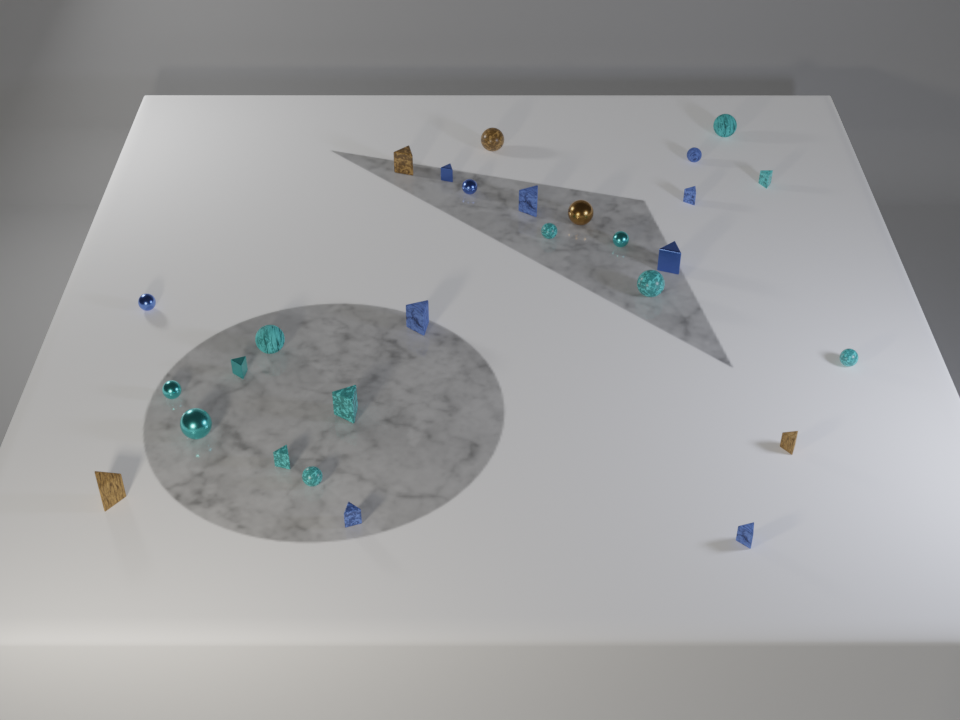}
    \begin{minipage}[t][2.2cm][t]{1\textwidth}
      \footnotesize
      \textbf{Question:} On the gray marble plane where there are between 3 and 5 tiny cyan objects, is there any wooden triangular prism that has the same size as most metallic objects? \\*
      \textbf{Answer:}  False \\*
      \textbf{Quantifiers:} between, some (any), most \\*[6pt]
    \end{minipage}\\*
    \begin{minipage}[t][2.2cm][t]{1\textwidth}
      \footnotesize
      \textbf{Question:} On the gray plane where there are between 1 and 4 cyan triangular prisms, 2 to 5 cyan items are the same material as most small items; is it right? \\*
      \textbf{Answer:} True \\*
      \textbf{Quantifiers:} between, between, most \\*[6pt]
    \end{minipage}\\*
    \begin{minipage}[t][2.2cm][t]{1\textwidth}
      \footnotesize
      \textbf{Question:} On the planes where there are between 3 and 6 cyan items on each plane, are there exactly 3 small triangular prisms that have the same color as most small metallic spheres? \\*
      \textbf{Answer:} True \\*
      \textbf{Quantifiers:} each, between, exactly N, most  \\*[6pt]
    \end{minipage}
  \end{minipage}
  \hspace{3.5mm}
  \begin{minipage}{0.48\textwidth}
    \includegraphics[width=\textwidth]{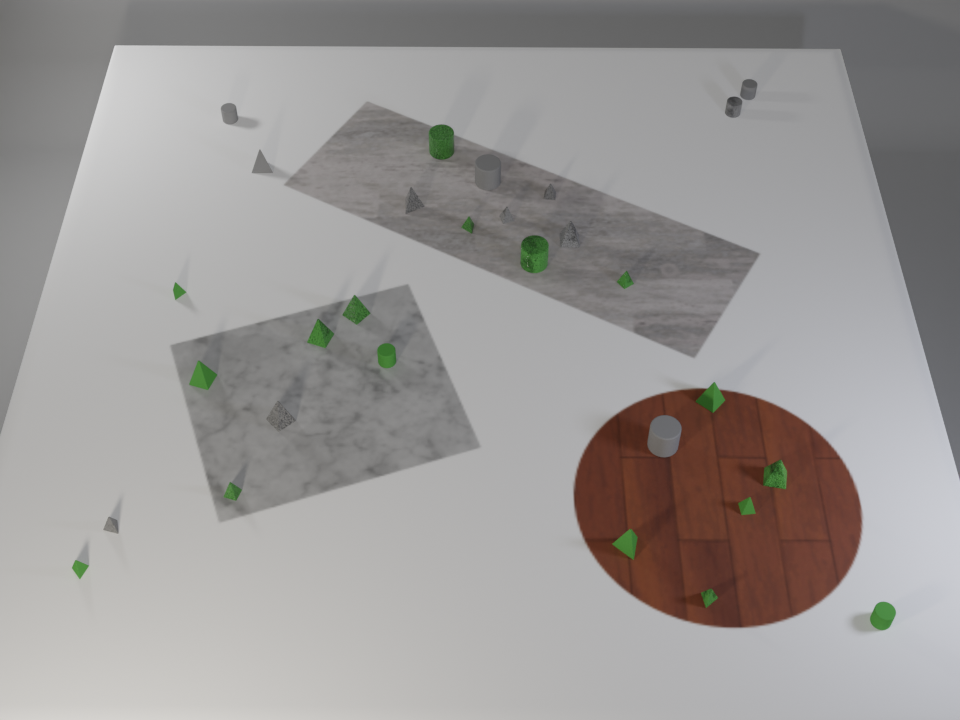}
    \begin{minipage}[t][2.2cm][t]{1\textwidth}
      \footnotesize
      \textbf{Question:} On the gray quadrilateral plane where there are not between 1 and 3 small gray items, are all the large items but at most 1 the same color as most leather items? \\*
      \textbf{Answer:} True \\*
      \textbf{Quantifiers:} not between, all but at most N, most \\*[6pt]
    \end{minipage}\\*
    \begin{minipage}[t][2.2cm][t]{1\textwidth}
      \footnotesize
      \textbf{Question:} On the plane where there are not between 2 and 4 green leather objects, fewer than a half of the gray cylinders are the same size as most gray rubbery objects; is it right? \\*
      \textbf{Answer:} False \\*
      \textbf{Quantifiers:} not between, fewer than F, most \\*[6pt]
    \end{minipage}\\*
    \begin{minipage}[t][2.2cm][t]{1\textwidth}
      \footnotesize
      \textbf{Question:} On the planes where there are not between 0 and 3 big objects on each plane, more than five twelfths of the big gray objects have the same shape as most gray rubber objects; is it right? \\*
      \textbf{Answer:} False \\*
      \textbf{Quantifiers:} each, not between, more than F, most \\*[6pt]
    \end{minipage}
  \end{minipage}
\end{figure*}

\begin{figure*}  
\begin{minipage}{0.48\textwidth}
    \includegraphics[width=\textwidth]{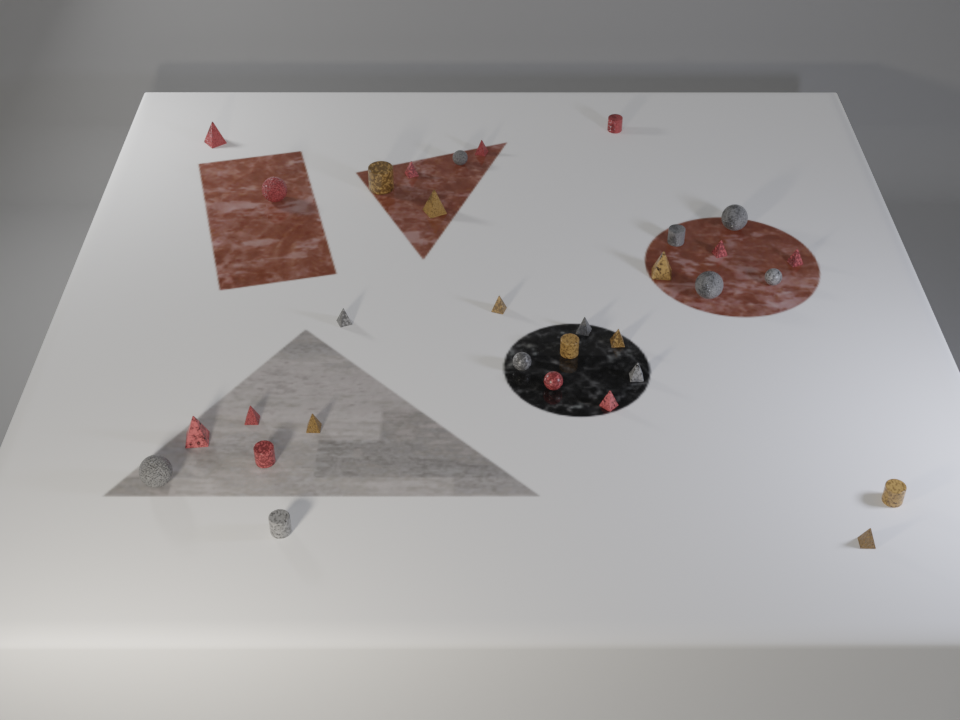}
    \begin{minipage}[t][2.2cm][t]{1\textwidth}
      \footnotesize
      \textbf{Question:} On the planes where there are not exactly 2 spheres on each plane, at most 4 red leather objects are the same shape as most objects; is it right? \\*
      \textbf{Answer:}  True \\*
      \textbf{Quantifiers:}  each, not exactly N, at most N, most\\*[6pt]
    \end{minipage}\\*
    \begin{minipage}[t][2.2cm][t]{1\textwidth}
      \footnotesize
      \textbf{Question:} On the triangular plane where there is not exactly 1 small marbled square-based pyramid, at least 2 red marbled items have the same size as most red items; is it right? \\*
      \textbf{Answer:} False \\*
      \textbf{Quantifiers:}  not exactly N, at least N, most\\*[6pt]
    \end{minipage}\\*
    \begin{minipage}[t][2.2cm][t]{1\textwidth}
      \footnotesize
      \textbf{Question:} On the marbled planes where there is at least 1 large marbled item on each plane, at least 2 large items are not the same material as most gray items; is it right? \\*
      \textbf{Answer:} True \\*
      \textbf{Quantifiers:} each, at least N, all but at least N (at least N $\neg$), most \\*[6pt]
    \end{minipage}
  \end{minipage}
  \hspace{3.5mm}
  \begin{minipage}{0.48\textwidth}
    \includegraphics[width=\textwidth]{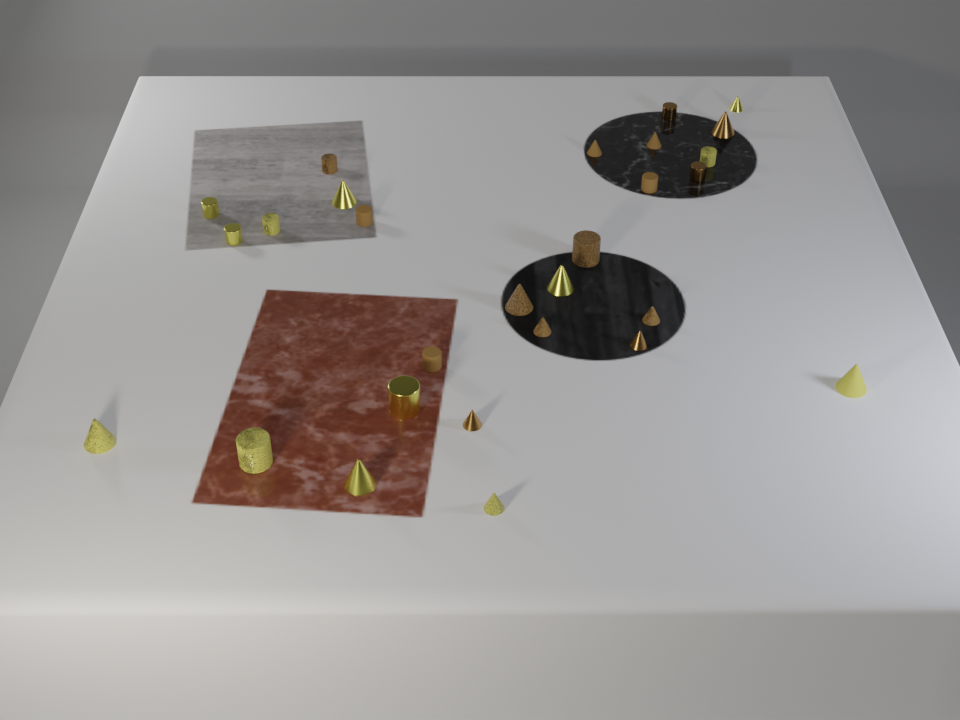}
    \begin{minipage}[t][2.2cm][t]{1\textwidth}
      \footnotesize
      \textbf{Question:}  On the planes where there are more than or equal to 2 cylinders on each plane, are there more brown metallic cones than brown leathery cylinders? \\*
      \textbf{Answer:} False \\*
      \textbf{Quantifiers:} each, at least N, more O$_1$ than O$_2$\\*[6pt]
    \end{minipage}\\*
    \begin{minipage}[t][2.2cm][t]{1\textwidth}
      \footnotesize
      \textbf{Question:}  On the wood plane where there is not exactly 1 small yellow leathery object, is the number of small objects less than the number of brown leathery cones? \\*
      \textbf{Answer:}  False \\*
      \textbf{Quantifiers:}  not exactly N, fewer O$_1$ than O$_2$ \\*[6pt]
    \end{minipage}\\*
    \begin{minipage}[t][2.2cm][t]{1\textwidth}
      \footnotesize
      \textbf{Question:} On the planes where there are no fewer than 2 big cones on each plane, is the number of circular cylinders the same as the number of big circular cylinders? \\*
      \textbf{Answer:} True \\*
      \textbf{Quantifiers:} each, at least N, equal O$_1$ and O$_2$ \\*[6pt]
    \end{minipage}
  \end{minipage}
  \vspace{0.2cm}
  
  \begin{minipage}{0.48\textwidth}
    \includegraphics[width=\textwidth]{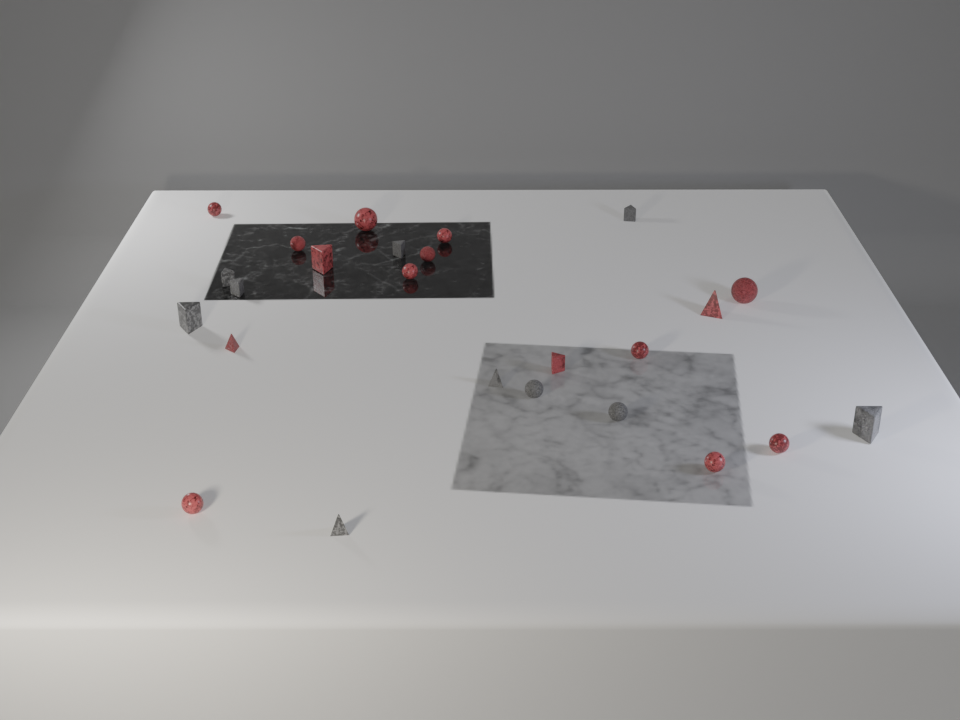}
    \begin{minipage}[t][2.2cm][t]{1\textwidth}
      \footnotesize
      \textbf{Question:} On the dappled planes, is there the same number of tiny objects on the left side of the big sphere and tiny red spheres right of the red leathery triangular prism? \\*
      \textbf{Answer:}  False \\*
      \textbf{Quantifiers:} equal O$_1$ and O$_2$ \\*[6pt]
    \end{minipage}\\*
    \begin{minipage}[t][2.2cm][t]{1\textwidth}
      \footnotesize
      \textbf{Question:} On the white non-geometric plane, are there fewer big objects to the right rear of the big red dappled object than tiny spheres in front of the big red dappled object? \\*
      \textbf{Answer:} True \\*
      \textbf{Quantifiers:} fewer O$_1$ than O$_2$ \\*[6pt]
    \end{minipage}\\*
    \begin{minipage}[t][2.2cm][t]{1\textwidth}
      \footnotesize
      \textbf{Question:} On the planes, is the number of triangular prisms to the right front of the red leathery tetrahedron greater than the number of leathery tetrahedrons on the left side of the large marble sphere? \\*
      \textbf{Answer:} True \\*
      \textbf{Quantifiers:} more O$_1$ than O$_2$  \\*[6pt]
    \end{minipage}
  \end{minipage}
  \hspace{3.5mm}
  \begin{minipage}{0.48\textwidth}
    \includegraphics[width=\textwidth]{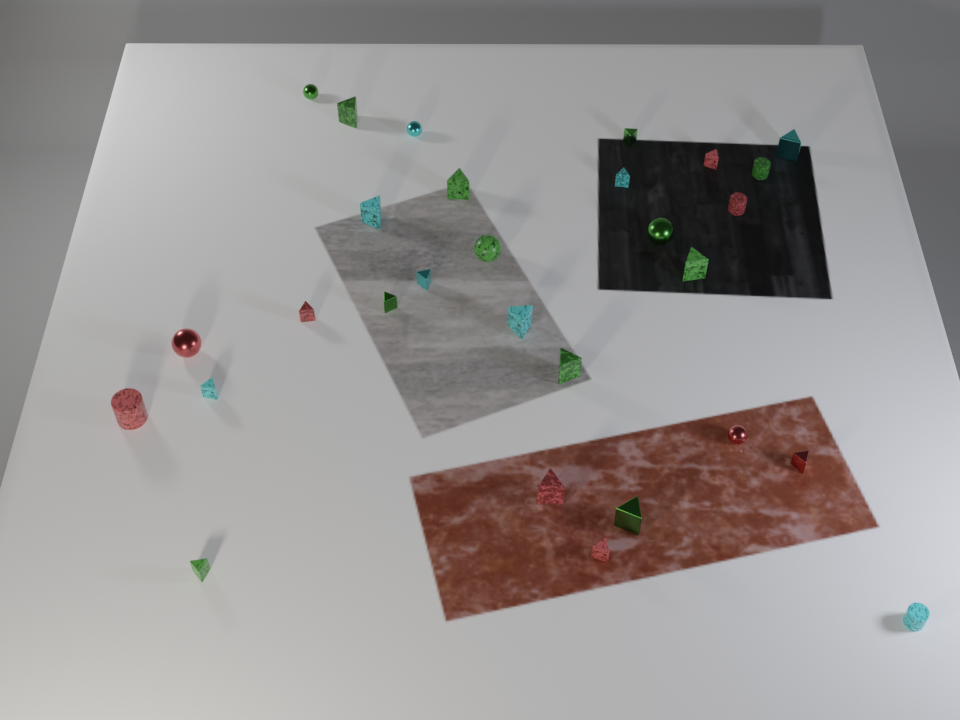}
    \begin{minipage}[t][2.2cm][t]{1\textwidth}
      \footnotesize
      \textbf{Question:} On the planes where there are no more than 3 tiny marbled objects on each plane, all the tiny marbled objects are in front of the red metallic triangular prism; is it right? \\*
      \textbf{Answer:} True \\*
      \textbf{Quantifiers:} each, at most N, all \\*[6pt]
    \end{minipage}\\*
    \begin{minipage}[t][2.2cm][t]{1\textwidth}
      \footnotesize
      \textbf{Question:} On the plane where there are at most 4 marble items, is there any tiny ball in front of the tiny red metal three-sided prism? \\*
      \textbf{Answer:} False \\*
      \textbf{Quantifiers:} at most N, some \\*[6pt]
    \end{minipage}\\*
    \begin{minipage}[t][2.2cm][t]{1\textwidth}
      \footnotesize
      \textbf{Question:} On the wood plane where there are fewer than or equal to 3 small marble objects, it is not the case that no large marble triangular prism is not on the left side of the green sphere; is it right? \\*
      \textbf{Answer:} True \\*
      \textbf{Quantifiers:} at most N, not all ($\neg$ no $\neg$) \\*[6pt]
    \end{minipage}
  \end{minipage}
\end{figure*}

\begin{figure*}  
\begin{minipage}{0.48\textwidth}
    \includegraphics[width=\textwidth]{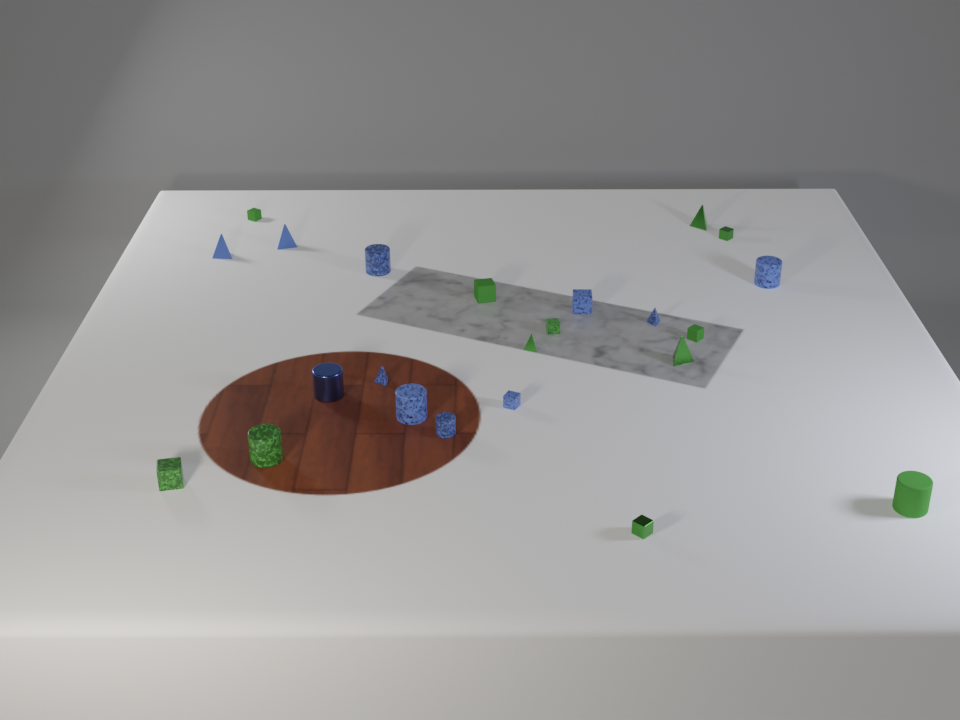}
    \begin{minipage}[t][2.2cm][t]{1\textwidth}
      \footnotesize
      \textbf{Question:} On the planes where there is exactly 1 small blue marbled object on each plane, every green metal object is not behind the green rubber cylinder; is it right? \\*
      \textbf{Answer:}  False \\*
      \textbf{Quantifiers:}  each, exactly N, no (every $\neg$)\\*[6pt]
    \end{minipage}\\*
    \begin{minipage}[t][2.2cm][t]{1\textwidth}
      \footnotesize
      \textbf{Question:} On the plane where there is not exactly 1 big green metal tetrahedron, some but not all of the marble objects are to the right of the blue tetrahedron; is it right? \\*
      \textbf{Answer:} True \\*
      \textbf{Quantifiers:}  not exactly N, some but not all\\*[6pt]
    \end{minipage}\\*
    \begin{minipage}[t][2.2cm][t]{1\textwidth}
      \footnotesize
      \textbf{Question:} On the planes where there is a total of 5 big green items, all the tiny green blocks but at most 1 are not right rear of the big green marble block; is it right? \\*
      \textbf{Answer:} False \\*
      \textbf{Quantifiers:} total, at most N (all but at most N $\neg$) \\*[6pt]
    \end{minipage}
  \end{minipage}
  \hspace{3.5mm}
  \begin{minipage}{0.48\textwidth}
    \includegraphics[width=\textwidth]{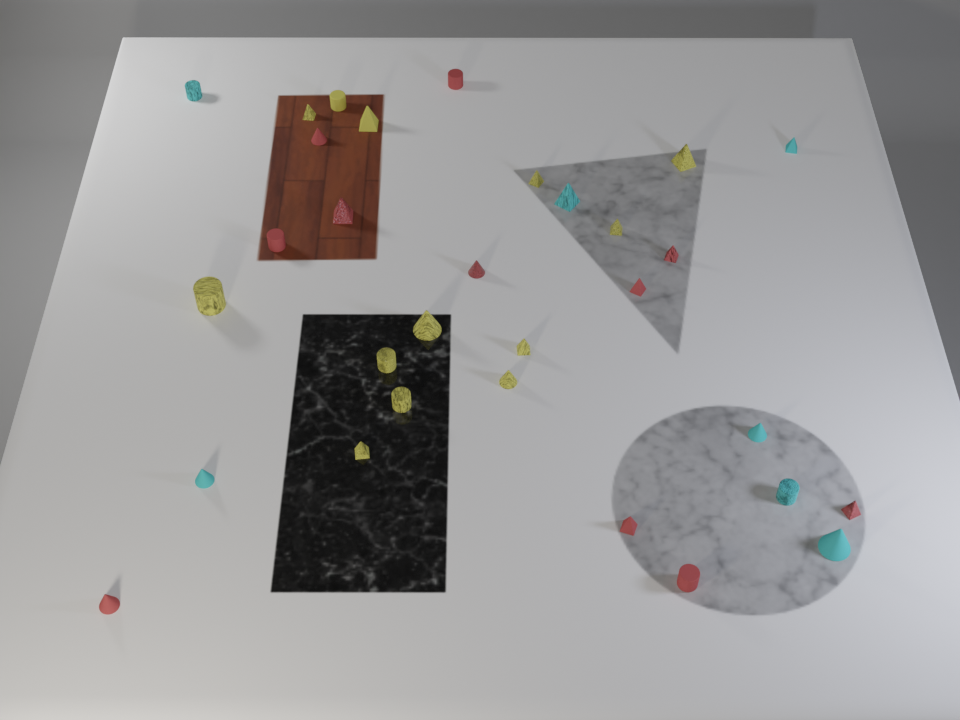}
    \begin{minipage}[t][2.2cm][t]{1\textwidth}
      \footnotesize
      \textbf{Question:}  On the planes where there are at least 2 cyan items on each plane, all the tiny cones but at least 4 are not right rear of the tiny red rubbery pentahedron; is it right? \\*
      \textbf{Answer:} False \\*
      \textbf{Quantifiers:} each, at least N, at least N (all but at least N $\neg$) \\*[6pt]
    \end{minipage}\\*
    \begin{minipage}[t][2.2cm][t]{1\textwidth}
      \footnotesize
      \textbf{Question:}  On the quadrilateral plane where there is at most 1 yellow circular cylinder, it is not the case that at most 1 red rubber object is left rear of the leather object; is it right? \\*
      \textbf{Answer:}  False \\*
      \textbf{Quantifiers:}  at most N, more than N ($\neg$ at most N) \\*[6pt]
    \end{minipage}\\*
    \begin{minipage}[t][2.2cm][t]{1\textwidth}
      \footnotesize
      \textbf{Question:} On the geometric plane whose material is different from that of other planes, it is not the case that all the red items but at least 2 are not right front of the tiny yellow cylinder; is it right? \\*
      \textbf{Answer:} True \\*
      \textbf{Quantifiers:} fewer than N ($\neg$ all but at least N $\neg$) \\*[6pt]
    \end{minipage}
  \end{minipage}
  \vspace{0.2cm}
  
  \begin{minipage}{0.48\textwidth}
    \includegraphics[width=\textwidth]{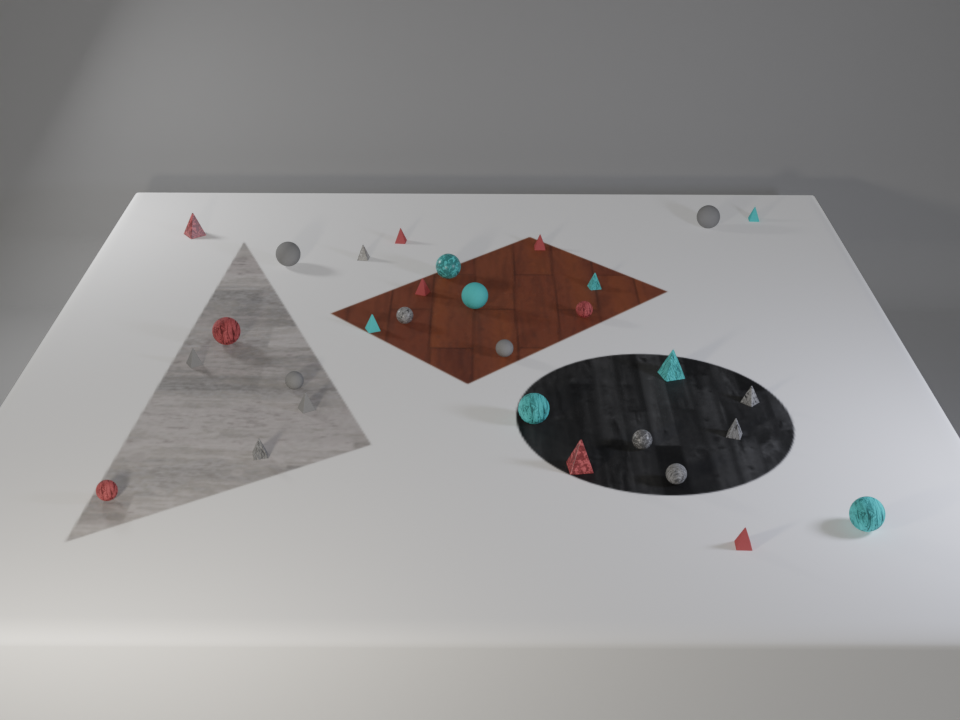}
    \begin{minipage}[t][2.2cm][t]{1\textwidth}
      \footnotesize
      \textbf{Question:} On the planes where there are not any small red rubber objects on each plane, there are exactly 2 small gray pentahedrons left rear of the large cyan pentahedron; is it right? \\*
      \textbf{Answer:}  False \\*
      \textbf{Quantifiers:} each, no ($\neg$ any), exactly N \\*[6pt]
    \end{minipage}\\*
    \begin{minipage}[t][2.2cm][t]{1\textwidth}
      \footnotesize
      \textbf{Question:} On the planes where there is not exactly 1 wood ball on each plane, are there 1 to 3 wood objects to the left front of the gray marbled square-based pyramid? \\*
      \textbf{Answer:} False \\*
      \textbf{Quantifiers:} each, not exactly N, between \\*[6pt]
    \end{minipage}\\*
    \begin{minipage}[t][2.2cm][t]{1\textwidth}
      \footnotesize
      \textbf{Question:} On the wooden planes where there are not between 1 and 3 red rubber items on each plane, at most 3 small items are not on the right front side of the big red sphere; is it right? \\*
      \textbf{Answer:} True \\*
      \textbf{Quantifiers:} each, not between, all but at most N (at most N $\neg$)  \\*[6pt]
    \end{minipage}
  \end{minipage}
  \hspace{3.5mm}
  \begin{minipage}{0.48\textwidth}
    \includegraphics[width=\textwidth]{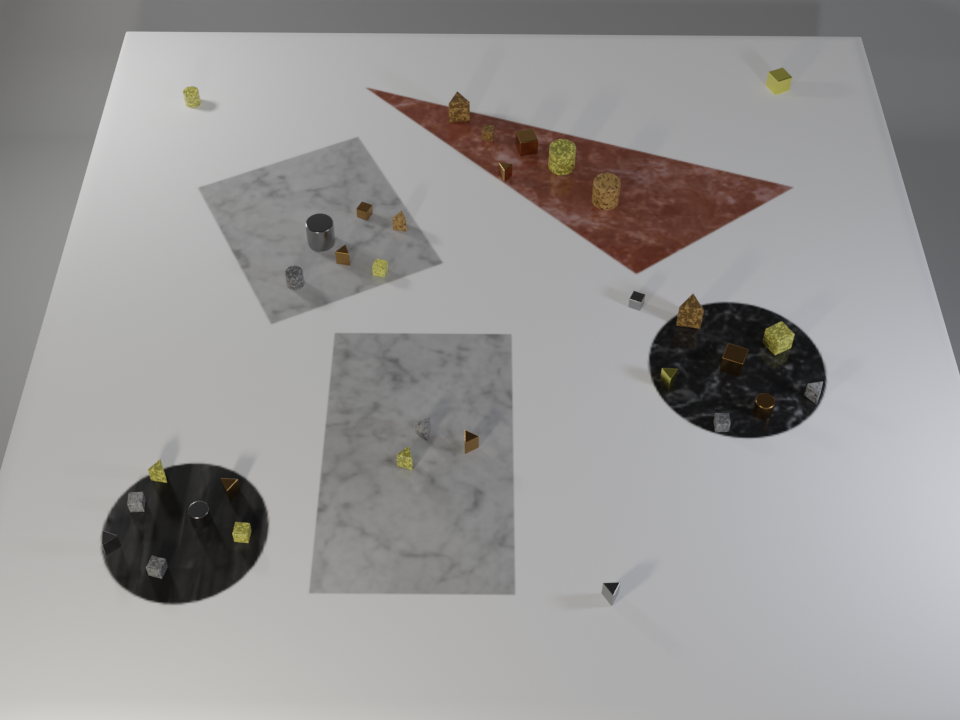}
    \begin{minipage}[t][2.2cm][t]{1\textwidth}
      \footnotesize
      \textbf{Question:} On the planes where there are between 1 and 3 tiny gray items on each plane, are all the tiny items but at least 5 right of the large gray metal circular cylinder? \\*
      \textbf{Answer:} False \\*
      \textbf{Quantifiers:} each, between, all but at least N \\*[6pt]
    \end{minipage}\\*
    \begin{minipage}[t][2.2cm][t]{1\textwidth}
      \footnotesize
      \textbf{Question:} On the round plane to the left of the brown dappled three-cornered plane, most items are behind the small yellow dappled block; is it right? \\*
      \textbf{Answer:} True \\*
      \textbf{Quantifiers:} most \\*[6pt]
    \end{minipage}\\*
    \begin{minipage}[t][2.2cm][t]{1\textwidth}
      \footnotesize
      \textbf{Question:} On the marble planes where there is a total of 7 tiny triangular prisms, it is not the case that fewer than 2/3 of the big items are to the left of the tiny gray cube; is it right? \\*
      \textbf{Answer:} False \\*
      \textbf{Quantifiers:} total, at least F ($\neg$ fewer than F) \\*[6pt]
    \end{minipage}
  \end{minipage}
\end{figure*}

\begin{figure*}  
\begin{minipage}{0.48\textwidth}
    \includegraphics[width=\textwidth]{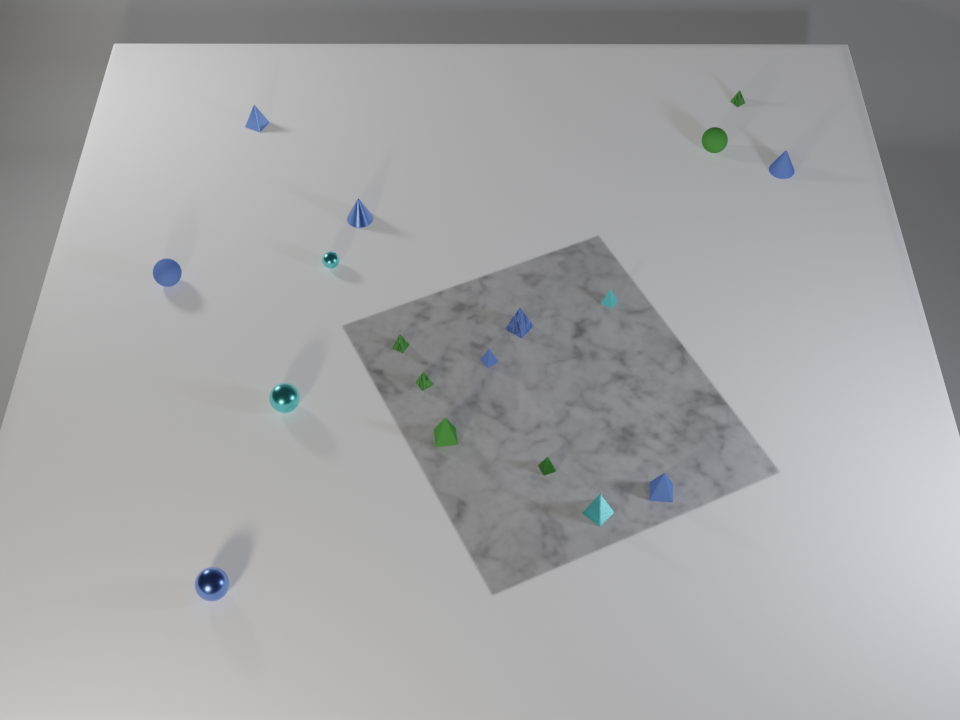}
    \begin{minipage}[t][2.2cm][t]{1\textwidth}
      \footnotesize
      \textbf{Question:} On the gray marble plane, it is not the case that more than 5/8 of the pentahedrons are to the right of the tiny metallic pentahedron; is it right? \\*
      \textbf{Answer:}  True \\*
      \textbf{Quantifiers:}  at most F ($\neg$ more than F) \\*[6pt]
    \end{minipage}\\*
    \begin{minipage}[t][2.2cm][t]{1\textwidth}
      \footnotesize
      \textbf{Question:} On the rectangular plane, all the blue wood square pyramids are larger than some rubber square pyramid; is it right? \\*
      \textbf{Answer:} True \\*
      \textbf{Quantifiers:}  all, some \\*[6pt]
    \end{minipage}\\*
    \begin{minipage}[t][2.2cm][t]{1\textwidth}
      \footnotesize
      \textbf{Question:} On the non-geometric plane, are all the green square-based pyramids smaller than some but not all of the square-based pyramids? \\*
      \textbf{Answer:} True \\*
      \textbf{Quantifiers:} all, some but not all \\*[6pt]
    \end{minipage}
  \end{minipage}
  \hspace{3.5mm}
  \begin{minipage}{0.48\textwidth}
    \includegraphics[width=\textwidth]{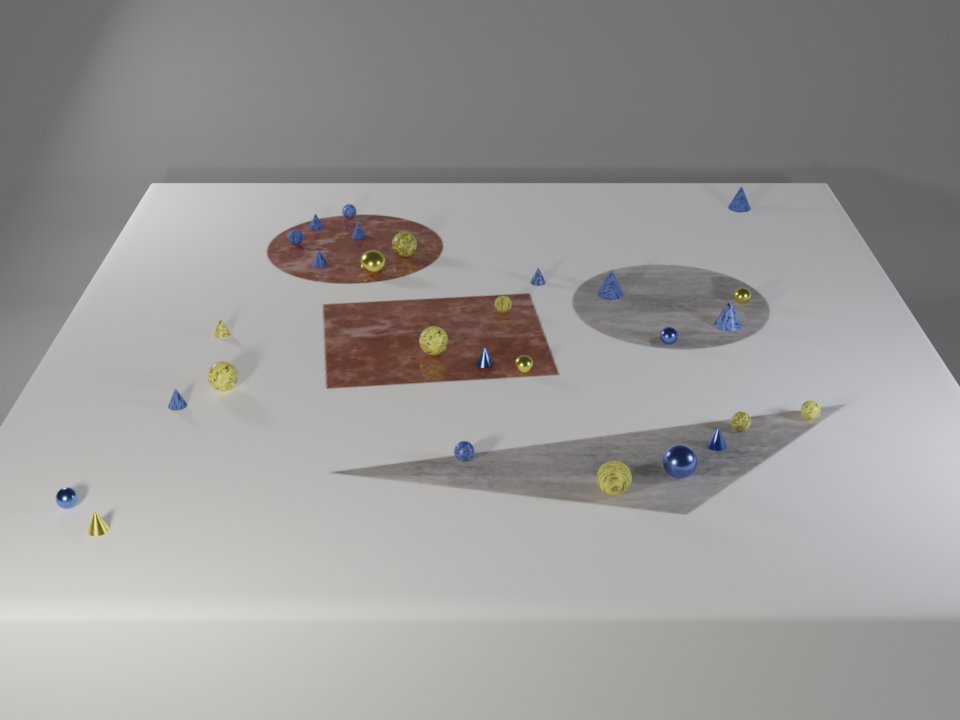}
    \begin{minipage}[t][2.2cm][t]{1\textwidth}
      \footnotesize
      \textbf{Question:}  On the planes where there are 0 large metal balls on each plane, it is not the case that no blue cone is larger than some but not all of the yellow cones; is it right? \\*
      \textbf{Answer:} False \\*
      \textbf{Quantifiers:} each, no (0), some ($\neg$ no), some but not all \\*[6pt]
    \end{minipage}\\*
    \begin{minipage}[t][2.2cm][t]{1\textwidth}
      \footnotesize
      \textbf{Question:}  On the planes where there are not exactly 3 tiny blue objects on each plane, exactly 3 blue cones are larger than at least 2 blue wood cones; is it right? \\*
      \textbf{Answer:}  False \\*
      \textbf{Quantifiers:} each, not exactly N, exactly N, at least N \\*[6pt]
    \end{minipage}\\*
    \begin{minipage}[t][2.2cm][t]{1\textwidth}
      \footnotesize
      \textbf{Question:} On the planes where there are not between 1 and 4 tiny marbled spheres on each plane, all the spheres but at least 2 are larger than at most 1 yellow sphere; is it right? \\*
      \textbf{Answer:} True \\*
      \textbf{Quantifiers:} each, not between, at least N (all but at least N $\neg$), more than N ($\neg$ at most) \\*[6pt]
    \end{minipage}
  \end{minipage}
  \vspace{0.2cm}
  
  \begin{minipage}{0.48\textwidth}
    \includegraphics[width=\textwidth]{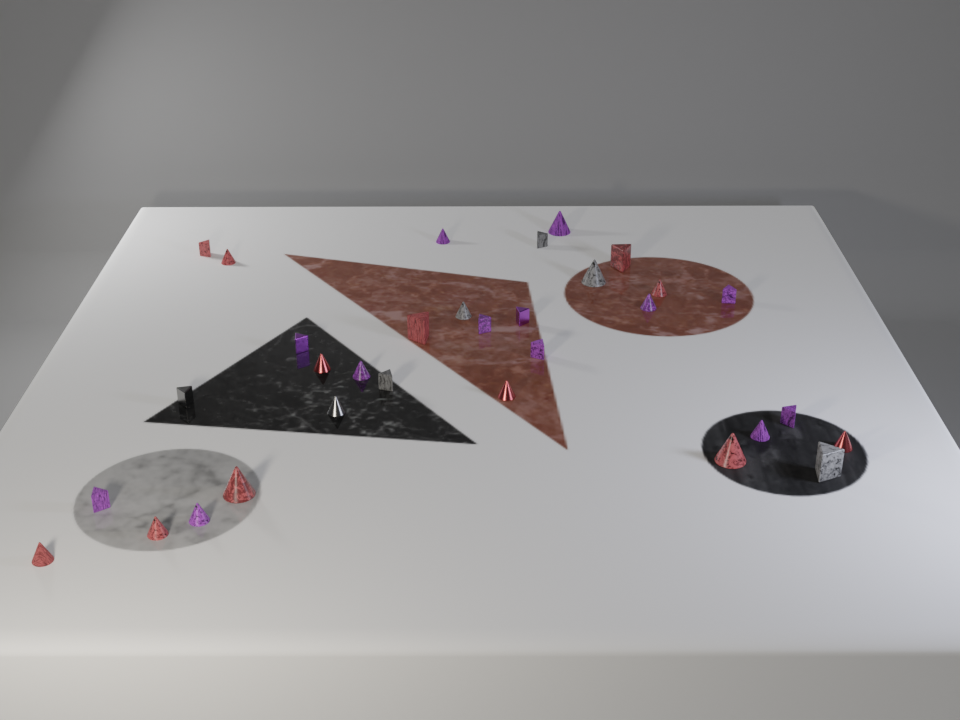}
    \begin{minipage}[t][2.2cm][t]{1\textwidth}
      \footnotesize
      \textbf{Question:} On the circular planes where there are exactly 2 big objects on each plane, 1 to 3 marbled cones are smaller than more than 3/7 of the red marbled cones; is it right? \\*
      \textbf{Answer:}  True \\*
      \textbf{Quantifiers:} each, exactly N, between, more than F \\*[6pt]
    \end{minipage}\\*
    \begin{minipage}[t][2.2cm][t]{1\textwidth}
      \footnotesize
      \textbf{Question:} On the brown plane where there are between 1 and 3 red marbled objects, at least one-ninth of the marbled cones are larger than at least 2 cones; is it right? \\*
      \textbf{Answer:} True \\*
      \textbf{Quantifiers:} between, at least N, at least F \\*[6pt]
    \end{minipage}\\*
    \begin{minipage}[t][2.2cm][t]{1\textwidth}
      \footnotesize
      \textbf{Question:} On the marble planes where there is a total of 5 marble items, all the cones but at least 2 are smaller than fewer than 3/4 of the gray cones; is it right? \\*
      \textbf{Answer:} False \\*
      \textbf{Quantifiers:} total, at least N (all but at least N $\neg$), at least F ($\neg$ fewer than F)\\*[6pt]
    \end{minipage}
  \end{minipage}
  \hspace{3.5mm}
  \begin{minipage}{0.48\textwidth}
    \includegraphics[width=\textwidth]{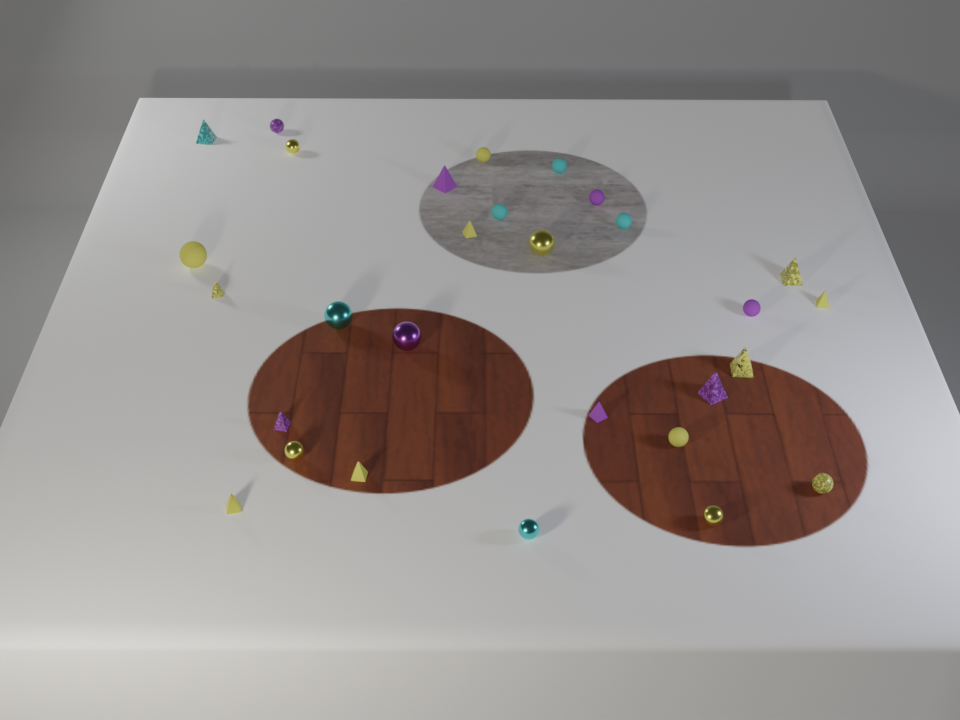}
    \begin{minipage}[t][2.2cm][t]{1\textwidth}
      \footnotesize
      \textbf{Question:} On the geometric plane whose color is different from that of other planes, all yellow spheres but 1 are smaller than fewer than 1 or more than 3 spheres; is it right? \\*
      \textbf{Answer:} True \\*
      \textbf{Quantifiers:} exactly N (all but N $\neg$), between ($\neg$ between) \\*[6pt]
    \end{minipage}\\*
    \begin{minipage}[t][2.2cm][t]{1\textwidth}
      \footnotesize
      \textbf{Question:} On the wooden planes where there are no fewer than 2 tiny balls on each plane, every item except the marble ball is not a tiny item; is it right? \\*
      \textbf{Answer:} False \\*
      \textbf{Quantifiers:} each, at least N, no \_ except (every \_ except $\neg$) \\*[6pt]
    \end{minipage}\\*
    \begin{minipage}[t][2.2cm][t]{1\textwidth}
      \footnotesize
      \textbf{Question:} On the brown plane where there is no more than 1 dappled square pyramid, no objects except the metal ones are not square pyramids; is it right? \\*
      \textbf{Answer:} True \\*
      \textbf{Quantifiers:} at most N, every \_ except (no \_ except $\neg$) \\*[6pt]
    \end{minipage}
  \end{minipage}
\end{figure*}

\end{document}